\documentclass[journal]{IEEEtran}

\usepackage[utf8]{inputenc} 
\usepackage[T1]{fontenc}    
\usepackage{hyperref}       
\usepackage{url}            
\usepackage{booktabs}       
\usepackage{amsfonts}       
\usepackage{nicefrac}       
\usepackage{microtype}      
\usepackage[caption=false,font=footnotesize]{subfig}
\usepackage{enumitem}
\usepackage{mathtools}
\usepackage{standalone} 
\usepackage{pgfplots}
\usepackage{todonotes}
\usepackage{bbm}
\usetikzlibrary{arrows,positioning,automata,bayesnet}
\usepgfplotslibrary{external}
\graphicspath{{./graphics/}}

\allowdisplaybreaks

\begin{document}

\title{Spatio-Temporal Structured Sparse Regression with Hierarchical Gaussian Process Priors}

\author{Danil~Kuzin,
	Olga~Isupova,
        and~Lyudmila~Mihaylova,~\IEEEmembership{Senior Member,~IEEE}
\thanks{D.Kuzin, L.Mihaylova are with the Department of Automatic Control and Systems Engineering, the University of Sheffield, Sheffield,
UK e-mail: dkuzin1@sheffield.ac.uk, l.s.mihaylova@sheffield.ac.uk. O.Isupova is with the Department of Engineering Science, the University of Oxford, Oxford, UK e-mail: olga.isupova@eng.ox.ac.uk}}

\markboth{IEEE Transactions on Signal Processing}%
{Shell \MakeLowercase{\textit{et al.}}: Spatio-temporal structured sparse modelling with hierarchical Gaussian process priors}

\maketitle

\begin{abstract}
This paper introduces a new sparse spatio-temporal structured Gaussian process regression framework for online and offline Bayesian inference. This is the first framework that gives a time-evolving representation of the interdependencies between the components of the sparse signal of interest. A hierarchical Gaussian process describes such structure and the interdependencies are represented via the covariance matrices of the prior distributions. The inference is based on the expectation propagation method and the theoretical derivation of the posterior distribution is provided in the paper. The inference framework is thoroughly evaluated over synthetic, real video and electroencephalography (EEG) data where the spatio-temporal evolving patterns need to be reconstructed with high accuracy. It is shown that it achieves 15\% improvement of the F-measure compared with the alternating direction method of multipliers, spatio-temporal sparse Bayesian learning method and one-level Gaussian process model. Additionally, the required memory for the proposed algorithm is less than in the one-level Gaussian process model. This structured sparse regression framework is of broad applicability to source localisation and object detection problems with sparse signals.

\end{abstract}

\section{Introduction}
\IEEEPARstart{S}{parse} regression problems arise often in various applications, e.g., compressive sensing~\cite{duarte2011structured}, EEG source localisation~\cite{gorodnitsky1997sparse} and direction of arrival estimation~\cite{yin2011direction}. In all these applications, a dictionary of basis functions can be constructed that allows sparse representations of the signals of interest, i.e. many of the coefficients of the basis functions are close to zero. This allows to perform sensing tasks with lower amount of observations than the signal dimensionality. However, the signal recovery problem becomes more computationally expensive when sparsity assumptions are incorporated.

The sparse signal representation can be expressed as a regression problem of finding a signal $\mathbf{x}$ given the vector $\mathbf{y}$ of observations and the design matrix $\mathbf{A}$ that satisfies the equation
\begin{equation}
\label{eq:lin_reg}
\mathbf{y}=\mathbf{A}\mathbf{x}+\boldsymbol\varepsilon,
\end{equation}
where $\boldsymbol\varepsilon$ is the Gaussian noise vector, $\boldsymbol\varepsilon \sim \mathcal{N}(\boldsymbol\varepsilon; \mathbf{0},\sigma^2\mathbf{I})$, $\sigma^2$ is the variance and $\mathbf{I}$ is the identity matrix. Therefore, the observations also have a Gaussian distribution
\begin{equation}
\mathbf{y} \sim \mathcal{N}(\mathbf{y}; \mathbf{A}\mathbf{x},\sigma^2\mathbf{I}).
\end{equation}

When the number of observations is less than the number of coefficients the problem is ill-posed in the sense that it has an infinite number of possible solutions and additional regularisation is required. This is usually achieved by imposing~$l_p$ penalty functions with~$0\le p < 2$~\cite{Tibshirani96,malioutov2005sparse,carmi2010methods}.

In the compressive sensing literature, it has been shown that if a matrix $\mathbf{A}$ satisfies the restricted isometry property (RIP)~\cite{candes2005decoding} then a solution of a convex $l_1$-minimisation problem is equivalent to a solution of a sparse $l_0$-minimisation problem. However, the problem of identification whether a given matrix satisfies the RIP is NP-hard~\cite{tillmann2014computational}. In contrast, Bayesian models do not impose any restrictions on the matrix $\mathbf{A}$ and regularise the problem (\ref{eq:lin_reg}) with sparsity-inducing priors~\cite{Tipping2001}.

Bayesian models for sparse regression can be classified into models with a \textit{weak sparsity} prior and a \textit{strong sparsity} prior~\cite{mohamed2011bayesian}. The weak sparsity prior leads to a unimodal posterior distribution of the signal with a sharp peak at zero, thus each coefficient has a high posterior probability of being close to zero. The strong sparsity prior is a mixture of latent binary variables that explicitly capture whether coefficients are zero or non-zero. In this paper we consider one type of strong sparsity priors --- spike and slab models.

In spike and slab models, sparsity is achieved by selecting each component of $\mathbf{x}$ from a mixture of a spike distribution, that is the delta function, and a slab distribution, that is some flat distribution, usually a Gaussian with a large variance~\cite{MitchellBeauchamp1988}. Following the Bayesian approach, latent variables that are indicators of spikes are added to the model~\cite{PolsonScott2010} and a relevant distribution is placed over them~\cite{murphy2012machine}. Therefore, each signal component has an independent latent variable, which controls whether this component would be a spike or a slab.

In many applications, the independence assumption is not valid \cite{bach2012structured} as non-zero elements tend to appear in groups, and an unknown structure often exists in the field of the latent variables. For example, wavelet coefficients of images are usually organised in trees~\cite{Mallat2008}, chromosomes have a spatial structure along a genome~\cite{hastie2015statistical}, video from single-pixel cameras has a temporal structure~\cite{Carin2014}. In these cases it is useful to introduce additional hierarchical or group penalties that promote such structures in recovered signals.

\IEEEpubidadjcol
\subsection{Contributions}
This paper proposes the spike and slab model with a hierarchical Gaussian process prior on the latent variables. Such hierarchical prior allows to model spatial structural dependencies for signal components that can evolve in time.

The model has a flexible structure which is governed only by the covariance functions of the Gaussian processes. This allows to model different types of structures and does not require any specific knowledge about the structure such as determination of particular groups of coefficients with similar behaviour. If, however, there is information about the structure, it can be easily incorporated into the covariance functions. The model is flexible as spatial and temporal dependencies are decoupled by different levels of the hierarchical Gaussian process prior. Therefore, the spatial and temporal structures are modelled independently allowing to encode different assumptions for each type of structure. It allows to reduce complexity and process streaming data.

Overall, the main contributions of this work consist in:
\begin{enumerate}
\item the proposed novel spike and slab model with the hierarchical Gaussian process prior for signal recovery with spatio-temporal structural dependencies;

\item the developed Bayesian inference algorithm based on expectation propagation;

\item the novel online inference algorithm for streaming data based on Bayesian filtering;

\item a thorough validation and evaluation of the proposed method over synthetic and real data including the electrical activity data for the EEG source localisation problem and video data for the compressive background subtraction problem.
\end{enumerate}

The paper is organised as follows. Section~\ref{sec:related_work} reviews the related work. Section~\ref{sec:review} provides an overview of existing spike and slab models. The proposed model and the inference algorithm are presented in Section~\ref{sec:model}. Section~\ref{sec:online_inference} demonstrates the online version of the algorithm. Section~\ref{sec:experiments} presents the complexity evaluation and numerical experiments. Section~\ref{sec:conclusions} concludes the paper. Appendices provide theoretical derivations of the inference algorithm.

\section{Related work}
\label{sec:related_work}
Different spatial structure assumptions for sparse models have been extensively studied in the literature. The group lasso~\cite{yuan2006model, sprechmann2011c} extends the classical lasso method for group sparsity such that coefficients form groups and all coefficients in a group are either non-zero or zero together, but groups are required to be defined in advance. In contrast to group lasso, structural dependencies in our model are defined by the parameters of covariance functions of the Gaussian processes (GPs) and the actual groups are inferred from the data.

Group constraints for weak sparse models include smooth relevance vector machines~\cite{schmolck2008smooth}, spatio-temporal coupling of the parameters for the scale mixture of Gaussians representation~\cite{van2010efficient, wu2014sparse}, row and element sparsity~\cite{chen2016simultaneous}, block sparsity~\cite{zhang2011sparse}.

For spike and slab priors a spatio-temporal structure is modelled with a one-level Gaussian processes prior~\cite{andersen2015bayesian}, where the prior is imposed on all locations of non-zero components together. The covariance matrix is represented as the Kronecker product of the temporal and spatial matrices.

In contrast to the one-level GP our model introduces an additional level of a GP prior for temporal dependencies. Therefore, the temporal and spatial structures are decoupled. The proposed model is thus more flexible. Broadly speaking, the top-level GP can encode the slow change of groups of spikes positions in time while the low-level GP allows to model the local changes of each group. The one-level GP prior model also requires significantly more memory to store the covariance function for modelling both spatial and temporal structural dependencies as it is built as a Kronecker product of spatial and temporal covariance matrices. The resulting size of the covariance matrix scales quadratically with spatio-temporal dimensionality, which makes it infeasible even for average size problems, whereas for our model the total size of two covariance matrices scales linearly.

More importantly, in the proposed model structural dependencies are considered at every timestamp whereas in~\cite{andersen2015bayesian} the GP prior is imposed on the whole batch of data. This consideration of every timestamp allows us to develop an incremental inference algorithm --- all latent variables are inferred for the new time moment in the similar manner as for the offline inference. Meanwhile, it is unclear how to apply the one-level GP model to the incremental data without re-processing the previous data.

GPs are widely used to model complex structures and dynamics in data not only in sparse problems. In~\cite{deisenroth2012expectation} GP is used as a prior for nonlinear state transition and observation functions for state-space Bayesian filtering. Hierarchical GP models are proposed to model structures in~\cite{lawrence2007hierarchical}.

\section{Sparse models for structured data}
\label{sec:review}
This section presents a roadmap of models that are used in the formulation of the proposed spatio-temporal structured sparse model. It starts from the basic spike and slab model and continues with its extension for structured data.

The generative model for the spatio-temporal regression problem can be formulated in the following way:
\begin{itemize}
\item The data is collected for the sequence of the $T$ discrete timestamps. Indexes are denoted by $t\in [1, \ldots, T]$.
\item At each timestamp $t$ the unknown signal of size $N$ is denoted by $\mathbf{x}_t = [x_{1t}, \ldots, x_{Nt}]^\top$. Signals at all timestamps are concatenated into a matrix $\mathbf{X} = [\mathbf{x}_1, \ldots, \mathbf{x}_T]$.
\item The observations of size $K$ are denoted by $\mathbf{y}_{t} = [y_{1t}, \ldots, y_{Kt}]^\top$. They are obtained with the design matrix $\mathbf{A} \in \mathbb{R}^{K \times N}$. Observations at all timestamps are concatenated into matrix $\mathbf{Y} = [\mathbf{y}_1, \ldots, \mathbf{y}_T]$.
\item An independent Gaussian noise with the variance $\sigma^2$ is added to the observations.
\end{itemize}
The probabilistic model can be then expressed as
\begin{equation}
\label{eq:problem_formulation}
p(\mathbf{y}_t | \mathbf{x}_t) = \mathcal{N}(\mathbf{y}_t; \mathbf{A}\mathbf{x}_t, \sigma^2 \mathbf{I}) \quad \forall t.
\end{equation}
It is assumed that the dimensionality $K$ of observations $\mathbf{y}_{t}$ is less than the dimensionality $N$ of signals $\mathbf{x}_t$, therefore the problem of recovery of signal $\mathbf{x}_t$ from observations $\mathbf{y}_t$ is underdetermined and it can have an infinite number of solutions. Sparsity-inducing priors allow to specify additional constraints that lead to a unique optimal solution.

\subsection{Factor graphs}
For Bayesian models, factor graphs are used to visualise complex distributions~\cite{wainwright2008graphical} in a form of undirected graphical models. They are also important for the approximate inference method described in Section \ref{sec:model}.

The joint probability density function $p(\cdot)$ of latent variables~$\zeta_i$ can be factorised as a product of factors $\psi_C$ that are functions of a corresponding set of latent variables $\boldsymbol\zeta_C$
\begin{equation}
	\label{eq:factor_decomposition}
	p(\zeta_1, ... , \zeta_m) = \dfrac{1}{Z}\prod_{C}\psi_C(\boldsymbol\zeta_C),
\end{equation}
where $Z$ is a normalisation constant.
This factorisation can be represented as a bipartite graph with variable vertices corresponding to $\zeta_i$, factor vertices corresponding to~$\psi_C$ and edges connecting corresponding vertices.

The distribution of latent variables $\mathbf{x}_t$ in (\ref{eq:problem_formulation}) can be represented as a factor
\begin{equation}
\label{eq:factor_g}
g_t(\mathbf{x}_t) = \mathcal{N}(\mathbf{y}_t; \mathbf{A}\mathbf{x}_t, \sigma^2 \mathbf{I}).
\end{equation}

The factor graphs are used in this paper to visualise different spike and slab models. In \figurename~\ref{pic:ssBer}~--~\ref{pic:ssSP-TMP} circles represent variable vertices and small squares represent factor vertices.

\subsection{Spike and slab model}
\label{subsec:ssmodel}
Sparsity can be induced with the spike and slab model~\cite{george1993variable}, where additional latent variables $\boldsymbol\Omega=\{\omega_{it}\}_{t=1:T,\,i=1:N}$ indicate if signal components $x_{it}$ are zeros. This is represented as a mixture of a spike and a slab
\begin{equation}
\label{eq:factor_f}
p(x_{it} | \omega_{it}) = \omega_{it}\delta_0(x_{it})+(1-\omega_{it})\mathcal{N}(x_{it}; 0, \sigma_x^2),
\end{equation}
where spike $\delta_0(\cdot)$ is the delta function centered at zero, and slab is the Gaussian distribution with the variance $\sigma_{x}^2$. The conditional distributions $p(x_{it} | \omega_{it})$ are further denoted by factors $f_{it}(\omega_{it}, x_{it})$.

In this model $\{\omega_{it}\}_{i = 1:N}$ are considered conditionally independent given $\mathbf{x}_t$. The prior is imposed on the indicators
\begin{equation}
\label{eq:omega_normal_Bernoulli}
p(\omega_{it}) = \text{Ber}(\omega_{it}; z),
\end{equation}
where $\text{Ber}(\cdot; z)$ denotes a Bernoulli distribution with the success probability parameter $z$. The prior distributions $p(\omega_{it})$ are further denoted by $h^{\text{ind}}_{it}(\omega_{it})$. The problem (\ref{eq:factor_g})~--~(\ref{eq:omega_normal_Bernoulli}) can be solved independently for each $t$.

\begin{figure}[!t]
\centering
	\includegraphics{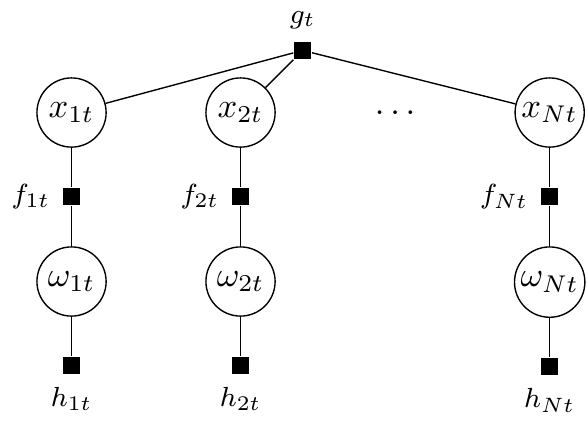}
\caption{Spike and slab model for one time moment (different time moments are independent). All signal components are conditionally independent given data, therefore structural assumptions cannot be modelled.}
\label{pic:ssBer}
\end{figure}

The model can be represented as a factor graph (\figurename~\ref{pic:ssBer}) with a product of factors (\ref{eq:factor_g})~--~(\ref{eq:omega_normal_Bernoulli}) for all $t$ and $i$.

The posterior $p(\mathbf{X}, \boldsymbol\Omega)$ of latent variables $\mathbf{X}$ and $\boldsymbol\Omega$ is
\begin{equation}
p=\prod\limits_{t = 1}^T \left[g_t(\mathbf{x}_t) \prod\limits_{i = 1}^N \left[f_{it}(\omega_{it}, x_{it}) h^{\text{ind}}_{it}(\omega_{it})\right]\right].
\end{equation}


\subsection{Spike and slab model with a spatial structure}
\label{subsec:gpmodel}
A spatial structure can be implemented by adding interdependencies for the locations of spikes in $x_{it}$~\cite{andersen2015bayesian, Engelhardt2014, wu2015high}. This is achieved by modelling the probabilities of spikes with the additional latent variables $\boldsymbol\Gamma=[\boldsymbol\gamma_{1}, \ldots, \boldsymbol\gamma_{T}]=\{\gamma_{it}\}_{t=1:T,\,i=1:N}$ that are samples from a Gaussian process. A Gaussian process is a way to specify prior on functions, it can be defined as an infinite expansion of multivariate Gaussian distribution. In GP all finite subsets of variables have a joint Gaussian distribution. The properties of the structure are defined through the covariance function of GP, which in this paper is assumed to be squared exponential:
\begin{equation}
\label{eq:gamma_factor}
p(\boldsymbol\gamma_t) = \mathcal{N}(\boldsymbol\gamma_t; \boldsymbol\mu_t, \boldsymbol\Sigma_0), \, \boldsymbol\Sigma_0(i,j) = \alpha_{\Sigma}\exp\left(-\dfrac{(i-j)^2}{2\ell^2_{\Sigma}}\right),
\end{equation}
where $\boldsymbol\mu_t$ is the mean vector and $\boldsymbol\Sigma_0$ is the covariance matrix with the hyperparameters $\alpha_{\Sigma}$ and $\ell^2_{\Sigma}$.

The conditional independence assumption for $\omega_{it}$ from~(\ref{eq:omega_normal_Bernoulli}) is replaced by
\begin{align}
\label{eq:omega_normal_Bernoulli_str}
p(\omega_{it} | \gamma_{it}) &= \text{Ber}(\omega_{it}; \Phi(\gamma_{it})), \\
\label{eq:factor_gamma}
p(\boldsymbol\gamma_{t}) &= \mathcal{N}(\boldsymbol\gamma_t; \boldsymbol\mu_t, \boldsymbol\Sigma_0),
\end{align}
where $\Phi(\cdot)$ is the standard Gaussian cumulative distribution function (cdf). Scaling is required to normalise probabilities to the $[0,1]$ interval and it is convenient to use $\Phi(\cdot)$ for this purpose in the derivations with GPs~\cite{rasmussen2006gaussian}. The conditional distributions $p(\omega_{it} | \gamma_{it})$ are denoted by factors $h_{it}(\omega_{it}, \gamma_{it})$. The prior distributions $p(\boldsymbol\gamma_{t})$ are denoted by $r^\text{ind}_t (\gamma_{t})$.

\begin{figure}[!t]
\centering
	\includegraphics{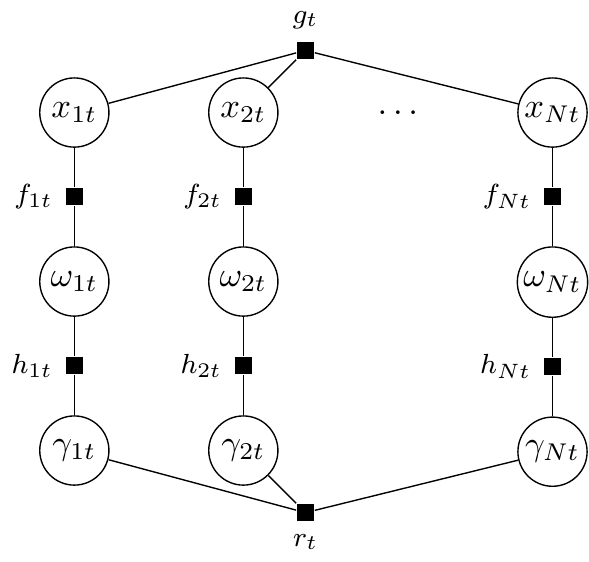}
\caption{Spike and slab model with a spatial structure for one time moment. The locations of spikes have a GP distribution, therefore encouraging a structure in space, but they are independent in time.}
\label{pic:ssGP}
\end{figure}

In this model $\{\boldsymbol\gamma_{t}\}_{t = 1:T}$ are independent and therefore the problem can be solved separately for each timestamp. Using the introduced factors (\ref{eq:factor_g}), (\ref{eq:factor_f}) and (\ref{eq:omega_normal_Bernoulli_str}) -- (\ref{eq:factor_gamma}), factor graph can be built as in Figure~\ref{pic:ssGP}. The posterior $p(\mathbf{X}, \boldsymbol\Omega, \boldsymbol\Gamma)$ of the latent variables is given by
\begin{equation}
p = \prod\limits_{t = 1}^T \left[ g_t(\mathbf{x}_t) \prod\limits_{i = 1}^N \left[f_{it}(\omega_{it}, x_{it}) h_{it}(\omega_{it}, \gamma_{it})\right] r_t(\boldsymbol\gamma_{t})\right].
\end{equation}


\section{The proposed spatio-temporal structured spike and slab model}
\label{sec:model}
In this paper a spatio-temporal latent structure of the positions of non-zero signal components is considered for the underdetermined recovery problem~(\ref{eq:problem_formulation}). The following assumptions are introduced:
\begin{enumerate}
\item $\mathbf{x}_t$ is sparse, i.e. it contains a lot of zeros for each timestamp $t$;
\item non-zero elements in $\mathbf{x}_t$ are clustered in groups for each timestamp $t$;
\item these groups can move and evolve in time.
\end{enumerate}

This recovery problem is addressed with the hierarchical Bayesian approach.
As in Section \ref{subsec:ssmodel}, the first assumption can be implemented in the model using the spike and slab prior~(\ref{eq:factor_f}).

Similarly to Section \ref{subsec:gpmodel}, the second model assumption can be implemented by adding spatial dependencies for the positions of spikes in $x_{it}$. This is achieved by modelling the probabilities of spikes $\boldsymbol\Omega$ with the scaled GP on $\boldsymbol\Gamma$~(\ref{eq:omega_normal_Bernoulli_str}),~(\ref{eq:factor_gamma}).
GPs specify a prior over an unknown structure. This is particularly useful as it allows to avoid a specification of any structural patterns --- the only parameter for structural modelling is the GP covariance function.

The third condition is addressed with the dynamic hierarchical GP prior. The mean $\mathbf{M} = \left[\boldsymbol\mu_1, \ldots, \boldsymbol\mu_T\right]$ for the spatial GP evolves over time according to the top-level temporal GP
\begin{equation}
\label{eq:model5}
\boldsymbol\mu_t \sim \mathcal{N} (\boldsymbol\mu_t; \boldsymbol\mu_{t-1}, \mathbf{W}), \, \mathbf{W}(i,j) = \alpha_{W}\exp\left(-\dfrac{(i-j)^2}{2\ell^2_W}\right),
\end{equation}
where $\mathbf{W}$ is the squared exponential covariance matrix of the temporal GP with the hyperparameters $\alpha_{W}$ and $\ell^2_W$.

This allows to implicitly specify the prior over the evolution function of the structure. The rate of the evolution is controlled with the top-level GP covariance function.

\begin{figure*}[t]
\centering
\includegraphics{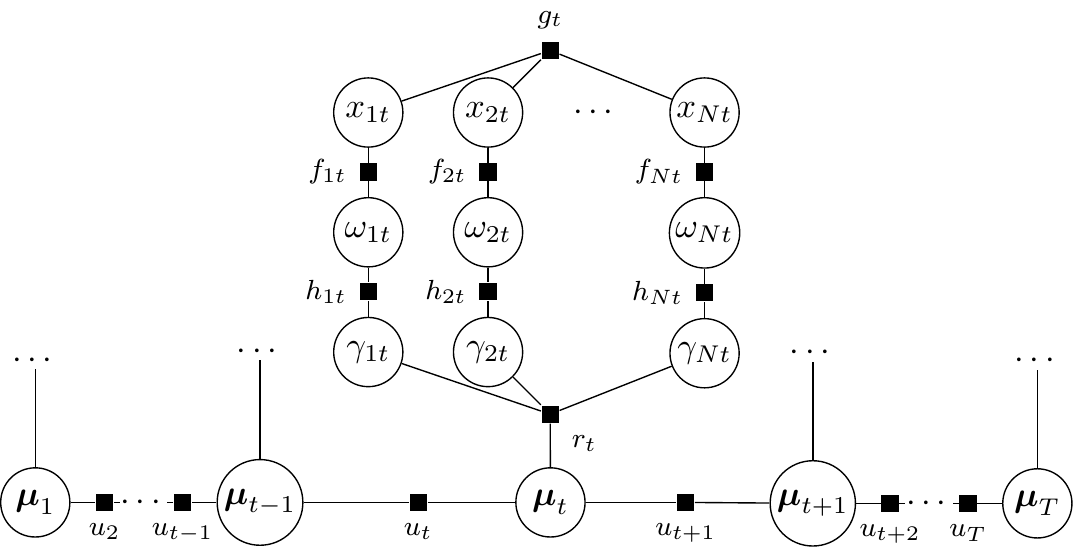}
\caption{Proposed spike and slab model with a spatio-temporal structure. The locations of spikes have a GP distribution in space with parameters that are controlled by a top-level GP and they evolve in time, therefore promoting temporal dependence.}
\label{pic:ssSP-TMP}
\end{figure*}

According to these assumptions, the model can be expressed as a factor graph (Figure~\ref{pic:ssSP-TMP}) where the factor $r_t(\boldsymbol\gamma_t, \boldsymbol\mu_t)$ denotes $\mathcal{N}(\boldsymbol\gamma_t; \boldsymbol\mu_t, \boldsymbol\Sigma_0)$ and the factor $u_t(\boldsymbol\mu_t, \boldsymbol\mu_{t-1})$ denotes $\mathcal{N} (\boldsymbol\mu_t; \boldsymbol\mu_{t-1}, \mathbf{W})$.

The full posterior distribution $p(\mathbf{X}, \boldsymbol\Omega, \boldsymbol\Gamma, \mathbf{M})$ is then
\setlength{\arraycolsep}{0.0em}
\begin{eqnarray}
\label{eq:P}
p&{}={}&\prod\limits_{t=1}^T \left[g_t(\mathbf{x}_t) \prod\limits_{i=1}^N \left[f_{it}(x_{it}, \omega_{it}) h_{it}(\omega_{it}, \gamma_{it})\right] r_t(\boldsymbol\gamma_t, \boldsymbol\mu_t)\right] \nonumber\\*
&&{\times}\: \prod\limits_{t = 2}^T u_t(\boldsymbol\mu_t, \boldsymbol\mu_{t-1}).
\end{eqnarray}
\setlength{\arraycolsep}{5pt}

The exact posterior for the proposed hierarchical spike and slab model is intractable, therefore approximate inference methods should be used. In this paper expectation propagation~(EP)~\cite{minka2001expectation} is employed. EP is shown to be the most effective Bayesian inference method for sparse modelling~\cite{hernandez2015expectation}.

In this section the description of the EP method and the key components of the inference for the proposed model are presented. The details of the inference algorithm can be found in the appendices.

\subsection{Expectation propagation}
\label{subsec:EP}
EP is a deterministic inference method that approximates the posterior distribution using the factor decomposition (\ref{eq:factor_decomposition}), where each factor is approximated with distributions $\tilde{\psi}_C(\cdot)$ from the exponential family:
\begin{equation}
	\tilde{p}(\zeta_1, ... , \zeta_m) = \frac{1}{\tilde{Z}}\prod_{C}\tilde{\psi}_C(\zeta_C),
\end{equation}
where $\tilde{p}$ is an approximating distribution and $\tilde{Z}$ is a normalisation constant. Approximating factorised distribution is determined by minimisation of the Kullback-Leibler~(KL) divergence with the true distribution. The KL-divergence is a common measure of similarity between distributions.

Direct approximation is intractable due to intractability of the true posterior. Minimisation of the KL divergence between individual factors $\psi_C$ and $\tilde{\psi}_C$ may not provide good approximation for the resulted product. In EP, approximation of each factor is performed in the context of other factors to improve a result for the final product.
Iteratively one of the factors is chosen for refinement. The chosen factor $\tilde{\psi}_C$ is refined to minimise the KL-divergence between the product $q \propto \tilde{\psi}_C \prod_{C' \neq C} \tilde{\psi}_{C'}$ and $\psi_C \prod_{C' \neq C} \tilde{\psi}_{C'}$, where the approximating factor is replaced with a factor from the true posterior.

Factor refinement consists of five steps which are summarised below (with details given in Appendices \ref{sec:update_f}-\ref{sec:update_u}).
\begin{enumerate}
\item \textit{Compute a cavity distribution} $q^{\setminus C} \propto \dfrac{q}{\tilde{\psi}_C}$: the joint distribution without the factor $\tilde{\psi}_C$
\item \textit{Compute a tilted distribution} $\psi_C q^{\setminus C}$: the product of the cavity distribution and the true factor
\item \textit{Refine the approximation} $q$: $q^* = \text{argmin } \text{KL}\left(\psi_C q^{\setminus C} || q\right)$ by minimising the KL-divergence between the tilted distribution $\psi_C q^{\setminus C}$ and the approximating distribution~$q$. This is equivalent to matching the moments of the distributions~\cite{minka2001expectation}.
\item \textit{Compute an updated factor} $\tilde{\psi}_C^{\text{new}} \propto \dfrac{q^*}{q^{\setminus C}}$ using the refined approximation and cavity distribution.
\label{item:update_factor}
\item \textit{Update the current joint posterior} $q^{\text{new}} \propto \tilde{\psi}_C^{\text{new}} \prod_{C' \neq C} \tilde{\psi}_{C'}$ with the newly updated factor $\tilde{\psi}_C^{\text{new}}$.
\end{enumerate}


\subsection{Approximating factors}
Here the key components of the EP inference algorithm for the proposed model are provided. The true posterior $p$~(\ref{eq:P}) is approximated with the distribution $q$
\begin{equation}
\label{eq:Q}
q = \prod_t q_{g_{t}} q_{f_{t}} q_{h_{t}} q_{r_{t}} q_{u_{t}},
\end{equation}
where each factor  $q_a$, $a \in \{g_t, f_t, h_t, r_t, u_t\}$, is from the exponential family and all latent variables are separated in the factors.

Below the factors $q_a$ of the approximating posterior $q$ are introduced. Gaussian and Bernoulli distributions are used in the factors, which parameters are updated during the iterations of the EP algorithm.

The factors $g_t =  \mathcal{N}(\mathbf{y}_t; \mathbf{A}\mathbf{x}_t, \sigma^2 \mathbf{I})$ from~(\ref{eq:factor_g}) can be viewed as the distributions of~$\mathbf{x}_t$ with fixed observed variables~$\mathbf{y}_t$: $q_{g_{t}} = \mathcal{N}(\mathbf{x}_{t}; \mathbf{m}_{g_{t}}, \mathbf{V}_{g_{t}})$, where
$\mathbf{m}_{g_{t}} = (\mathbf{A}^\top\mathbf{A})^{-1}\mathbf{A}^\top\mathbf{y}_t$,  $\mathbf{V}_{g_{t}} = \sigma^2 (\mathbf{A}^\top\mathbf{A})^{-1}$.

The factors $f_{t} = \prod_{i =1}^N f_{it}$ and $h_{t} = \prod_{i=1}^N h_{it}$ from~(\ref{eq:factor_f}) and~(\ref{eq:omega_normal_Bernoulli_str}) are approximated with the products of Gaussian and Bernoulli distributions
\begin{align}
q_{f_{t}} &= \mathcal{N}(\mathbf{x}_{t}; \mathbf{m}_{f_{t}}, \mathbf{V}_{f_{t}}) \prod_{i=1}^N \text{Ber}(\omega_{it}; \Phi(z_{f_{it}})),\\
q_{h_{t}} &= \mathcal{N}(\boldsymbol\gamma_{t}; \boldsymbol\nu_{h_{t}}, \mathbf{S}_{h}) \prod_{i=1}^N \text{Ber}(\omega_{it}; \Phi(z_{h_{it}})),
\end{align}
where the components of $\mathbf{x}_t$ and $\boldsymbol\gamma_t$ are independent. Therefore, the covariance matrices $\mathbf{V}_{f_{t}}$ and $\mathbf{S}_{h}$ are diagonal\footnote{Note that $\mathbf{S}_{h}$ does not depend on time. In this paper, single covariance matrices are used for all time moments for both GP variables $\boldsymbol\gamma$ and $\boldsymbol\mu$ in the approximating factors. However, the method can be applied with individual covariance matrices for each time moment as well.}. Distribution parameters $\mathbf{m}_{f_{t}}$, $\mathbf{V}_{f_{t}}$, $z_{f_{it}}$, $\boldsymbol\nu_{h_{t}}$, $\mathbf{S}_{h}$, and $z_{h_{it}}$ are updated during EP iterations according to Appendices \ref{sec:update_f} and \ref{sec:update_h}.

The approximation for the factors $r_{t} = \mathcal{N}(\boldsymbol\gamma_t; \boldsymbol\mu_t, \boldsymbol\Sigma_0)$ and $u_{t} = \mathcal{N} (\boldsymbol\mu_t; \boldsymbol\mu_{t-1}, \mathbf{W})$ from~(\ref{eq:gamma_factor}) and~(\ref{eq:model5}) is intended to separate the latent variables and it is represented as products of Gaussian distributions
\begin{align}
q_{r_{t}} &= \mathcal{N}(\boldsymbol\gamma_{t}; \boldsymbol\nu_{r_{t}}, \mathbf{S}_{r}) \mathcal{N}(\boldsymbol\mu_{t}; \mathbf{e}_{r_{t}}, \mathbf{D}_{r}),\\
q_{u_{t}} &= \mathcal{N}(\boldsymbol\mu_{t-1}; \mathbf{e}_{u_{t}\leftarrow}, \mathbf{D}_{u\leftarrow}) \mathcal{N}(\boldsymbol\mu_{t}; \mathbf{e}_{u_{t}\rightarrow}, \mathbf{D}_{u\rightarrow}).
\end{align}
Distribution parameters $\mathbf{e}_{r_{t}}$, $\mathbf{D}_{r}$, $\boldsymbol\nu_{r_{t}}$, $\mathbf{S}_{r}$, $\mathbf{e}_{u_{t}\leftarrow}$, $\mathbf{D}_{u\leftarrow}$, $\mathbf{e}_{u_{t}\rightarrow}$, and $\mathbf{D}_{u\rightarrow}$ are updated during EP iterations according to Appendices \ref{sec:update_r} and \ref{sec:update_u}.

The posterior approximation $q$ given by~(\ref{eq:Q}) thus contains the products of Gaussian and Bernoulli distributions that are equal to unnormalised Gaussian and Bernoulli distributions, respectively (Appendix \ref{sec:app_rules}). This can be conveniently expressed in terms of the natural parameters and $q$ can be represented in terms of distributions of the latent variables.

For $\mathbf{x}_t$ in $q$ this product property leads to the Gaussian distribution $\mathcal{N}(\mathbf{x}_t; \mathbf{m}_t, \mathbf{V}_t)$ with natural parameters
\begin{equation}
\mathbf{V}_t^{-1} = \mathbf{V}_{g_t}^{-1} + \mathbf{V}_{f_t}^{-1}, \, \mathbf{V}_t^{-1}\mathbf{m}_t = \mathbf{V}_{g_t}^{-1}\mathbf{m}_{g_t} + \mathbf{V}_{f_t}^{-1}\mathbf{m}_{f_t}.
\end{equation}

Similarly, $\boldsymbol\gamma_t$ in $q$ is distributed as $\mathcal{N}(\boldsymbol\gamma_t; \boldsymbol\nu_t, \mathbf{S})$, where natural parameters are
\begin{equation}
\mathbf{S}^{-1} = \mathbf{S}_{h}^{-1} + \mathbf{S}_{r}^{-1}, \, \mathbf{S}^{-1}\boldsymbol\nu_t = \mathbf{S}_{h}^{-1}\boldsymbol\nu_{h_t} + \mathbf{S}_{r}^{-1}\boldsymbol\nu_{r_t}.
\end{equation}

The top GP latent variables $\boldsymbol\mu_t$ have the Gaussian distributions $\mathcal{N}(\boldsymbol\mu_t; \mathbf{e}_t, \mathbf{D})$ with natural parameters
\begin{subequations}
\begin{align}
&\mathbf{D}^{-1} = \mathbf{D}_{r}^{-1} +  \mathbf{D}_{u\rightarrow}^{-1}\mathbbm{1}_{t > 1} + \mathbf{D}_{u\leftarrow}^{-1}\mathbbm{1}_{t<T},\\
&\mathbf{D}^{-1}\mathbf{e}_t = \mathbf{D}_{r}^{-1}\mathbf{e}_{r_t} +  \mathbf{D}_{u\rightarrow}^{-1}\mathbf{e}_{u_t\rightarrow}\mathbbm{1}_{t > 1} + \nonumber\\
&\quad\mathbf{D}_{u\leftarrow}^{-1}\mathbf{e}_{u_{t+1}\leftarrow}\mathbbm{1}_{t<T},
\end{align}
\end{subequations}
where $\mathbbm{1}$ is the indicator function.

The distributions for $\boldsymbol\omega_t$ are $\prod_{i=1}^N\text{Ber}( \omega_{it}; \Phi(z_{it}))$ with the parameters
\begin{equation}
z_{it} = \Phi^{-1}\left( \left[\dfrac{(1 - \Phi(z_{f_{it}}))(1 - \Phi(z_{h_{it}}))}{\Phi(z_{f_{it}}) \Phi(z_{h_{it}})} + 1 \right]^{-1} \right).
\end{equation}

The full approximating posterior~$q$ is then
\setlength{\arraycolsep}{0.0em}
\begin{eqnarray}
q&{}={}&\prod_{t=1}^T\mathcal{N}(\mathbf{x}_t; \mathbf{m}_t, \mathbf{V}_t) \prod_{t=1}^T\prod_{i=1}^N\text{Ber}( \omega_{it}; \Phi(z_{it})) \nonumber\\*
&&{\times}\: \prod_{t=1}^T\mathcal{N}(\boldsymbol\gamma_t; \boldsymbol\nu_t, \mathbf{S}) \prod_{t=1}^T\mathcal{N}(\boldsymbol\mu_t; \mathbf{e}_t, \mathbf{D}).
\end{eqnarray}
\setlength{\arraycolsep}{5pt}

In the EP inference algorithm, each of the introduced approximating factors $q_{f_{t}}$, $q_{h_{t}}$, $q_{r_{t}}$, $q_{u_{t}}$ is iteratively updated according to the factor refinement procedure as in Section~\ref{subsec:EP}. Note that the factors $q_{g_{t}}$ are not updated, as the corresponding factors $g_t$ from the true posterior distribution are already from the exponential family.

\subsection{Implementation details}
There are no theoretical guarantees of EP convergence. However, it can be achieved using \textit{damping}~\cite{minka2002expectation}: during step~\ref{item:update_factor} of the factor refinement procedure in Section~\ref{subsec:EP} the factor is updated as
$q_a^{\text{damp}} = (q_a^{\text{new}})^{\eta} (q_a^{\text{old}})^{1 - \eta}$,
where $q_a^{\text{old}}$ is the value of the factor from the previous iteration, $q_a^{\text{new}}$ is the updated value of the factor, $\eta \in (0, 1]$~is the damping coefficient. It is exponentially decreased as $\eta = \eta^\text{old} \xi$ after each iteration, where $\xi \in (0, 1]$ is the parameter that governs the speed of exponential decrease and $\eta^{\text{old}}$ is the value of the damping coefficient from the previous iteration.

It is also known that during the EP updates negative variances can appear~\cite{hernandez2015expectation}. In this case negative variances are replaced with a large value representing $+\infty$.

\section{Online Inference with Bayesian Filtering}
\label{sec:online_inference}
In this section the problem~(\ref{eq:problem_formulation}) is considered for streaming data, i.e. when new data becomes available at every timestamp. The conventional batch inference can be infeasible for large or streaming data. The developed online Bayesian filtering algorithm for the model presented in Section~\ref{sec:model} allows to iteratively update the approximation of $\mathbf{x}$ based on new samples of data.

Bayesian filtering consist of two steps that are iterated for each new sample of data:
\begin{itemize}
\item \textit{prediction}, where an estimate of a hidden system state at the next time step is predicted based on the observations available at the current time moment;
\item \textit{update}, where this estimate is updated once an observation at the next time moment is obtained.
\end{itemize}
In the proposed model the hidden state is represented by the latent variables $\mathbf{x}_t$, $\boldsymbol\omega_{t}$, $\boldsymbol\gamma_t$ and $\boldsymbol\mu_t$ that should be inferred based on observations~$\mathbf{y}_t$.

\subsection{Prediction}
At the prediction step for the timestamp $t+1$ the current estimate of the posterior distribution of the latent variables $p(\mathbf{x}_{t}, \boldsymbol\omega_{t}, \boldsymbol\gamma_{t}, \boldsymbol\mu_{t} | \mathbf{y}_{1:t})$ is available. It is based on all observations $\mathbf{y}_{1:t} = [\mathbf{y}_1, \ldots, \mathbf{y}_t]$ up to the timestamp $t$. The initial estimate of this posterior can be obtained by the offline inference algorithm applied to the initial $T_{\text{init}}$ timestamps.

Marginalisation of the latent variables for the current timestamp $t$ allows to obtain predictions for the latent variables for the next timestamp $t+1$
\begin{align}
p(\mathbf{x}_{t+1}&, \boldsymbol\omega_{t+1}, \boldsymbol\gamma_{t+1}, \boldsymbol\mu_{t+1} | \mathbf{y}_{1:t} ) = \nonumber\\
= &\int p(\mathbf{x}_{t+1}, \boldsymbol\omega_{t+1}, \boldsymbol\gamma_{t+1}, \boldsymbol\mu_{t+1} | \mathbf{x}_{t}, \boldsymbol\omega_{t}, \boldsymbol\gamma_{t}, \boldsymbol\mu_{t}) \nonumber \\
\label{eq:predict_int}
&{}\times p(\mathbf{x}_{t}, \boldsymbol\omega_{t}, \boldsymbol\gamma_{t}, \boldsymbol\mu_{t} | \mathbf{y}_{1:t}) \mathrm{d} \mathbf{x}_{t} \mathrm{d}\boldsymbol\omega_{t} \mathrm{d}\boldsymbol\gamma_{t} \mathrm{d}\boldsymbol\mu_{t}
\end{align}

The first term in the integral~(\ref{eq:predict_int}) is factorised according to the generative model~(\ref{eq:factor_g}),(\ref{eq:factor_f}),(\ref{eq:omega_normal_Bernoulli_str}), and (\ref{eq:model5})
\begin{align}
&p(\mathbf{x}_{t+1}, \boldsymbol\omega_{t+1}, \boldsymbol\gamma_{t+1}, \boldsymbol\mu_{t+1} | \mathbf{x}_{t}, \boldsymbol\omega_{t}, \boldsymbol\gamma_{t}, \boldsymbol\mu_{t})\nonumber \\
&= p(\mathbf{x}_{t+1} | \boldsymbol\omega_{t+1}) p(\boldsymbol\omega_{t+1} | \boldsymbol\gamma_{t+1}) p(\boldsymbol\gamma_{t+1} | \boldsymbol\mu_{t+1}) p(\boldsymbol\mu_{t+1} | \boldsymbol\mu_t)
\end{align}

Therefore, the terms related to variables $\mathbf{x}_{t+1}$, $\boldsymbol\omega_{t+1}$ and $\boldsymbol\gamma_{t+1}$ are independent from the integral variables in~(\ref{eq:predict_int}) and the integral can be rewritten as
\begin{align}
\int &p(\mathbf{x}_{t+1}, \boldsymbol\omega_{t+1}, \boldsymbol\gamma_{t+1}, \boldsymbol\mu_{t+1} | \mathbf{x}_{t}, \boldsymbol\omega_{t}, \boldsymbol\gamma_{t}, \boldsymbol\mu_{t}) \nonumber \\
&{}\times p(\mathbf{x}_{t}, \boldsymbol\omega_{t}, \boldsymbol\gamma_{t}, \boldsymbol\mu_{t} | \mathbf{y}_{1:t}) \mathrm{d} \mathbf{x}_{t} \mathrm{d}\boldsymbol\omega_{t} \mathrm{d}\boldsymbol\gamma_{t} \mathrm{d}\boldsymbol\mu_{t} \nonumber \\
= {} & p(\mathbf{x}_{t+1} | \boldsymbol\omega_{t+1}) p(\boldsymbol\omega_{t+1} | \boldsymbol\gamma_{t+1}) p(\boldsymbol\gamma_{t+1} | \boldsymbol\mu_{t+1}) \nonumber \\
\label{eq:predict_int_mu}
&\times \int p(\boldsymbol\mu_{t+1} | \boldsymbol\mu_t) p(\boldsymbol\mu_t | \mathbf{y}_{1:t}) \mathrm{d} \boldsymbol\mu_t
\end{align}

The initial estimate of the posterior $p(\boldsymbol\mu_{T_{\text{init}}} | \mathbf{y}_{1:T_{\text{init}}})$ obtained from the offline EP algorithm is a Gaussian distribution:
\begin{equation}
p(\boldsymbol\mu_{T_{\text{init}}} | \mathbf{y}_{1:T_{\text{init}}}) = \mathcal{N}(\boldsymbol\mu_{T_{\text{init}}} ; \mathbf{e}_{1:T_{\text{init}}}, \mathbf{D}_{1:T_{\text{init}}}),
\end{equation}
where $\mathbf{e}_{1:T_{\text{init}}}$ and $\mathbf{D}_{1:T_{\text{init}}}$ are the mean and the covariance matrix of the estimate of the posterior for $\boldsymbol\mu_{T_{\text{init}}}$ obtained based on observations $\mathbf{y}_{1:T_{\text{init}}}$.

According to the generative model~(\ref{eq:model5}) the first term of the integral in~(\ref{eq:predict_int_mu}) is also Gaussian, therefore the integral is also a Gaussian distribution on $\boldsymbol\mu_{t+1}$ for $t = T_{\text{init}}$:
\begin{align}
\int &p(\boldsymbol\mu_{t+1} | \boldsymbol\mu_t) p(\boldsymbol\mu_t | \mathbf{y}_{1:t}) \mathrm{d} \boldsymbol\mu_t \nonumber \\
\label{eq:mu_int}
= {} &\mathcal{N}(\boldsymbol\mu_{t+1} ; \mathbf{e}_{1:t}, \mathbf{D}_{1:t}^{\text{predict}}) \stackrel{\text{def}}{=} \hat{p}(\boldsymbol\mu_{t+1}),
\end{align}
where $\mathbf{D}_{1:t}^{\text{predict}} = \mathbf{W} + \mathbf{D}_{1:t}$ is the covariance of the predicted distribution.

Substitution of~(\ref{eq:predict_int_mu}) and~(\ref{eq:mu_int}) back into~(\ref{eq:predict_int}) provides the predicted distribution:
\begin{align}
p(&\mathbf{x}_{t+1}, \boldsymbol\omega_{t+1}, \boldsymbol\gamma_{t+1}, \boldsymbol\mu_{t+1} | \mathbf{y}_{1:t} )\nonumber\\
\label{eq:prediction}
= & p(\mathbf{x}_{t+1} | \boldsymbol\omega_{t+1}) p(\boldsymbol\omega_{t+1} | \boldsymbol\gamma_{t+1}) p(\boldsymbol\gamma_{t+1} | \boldsymbol\mu_{t+1}) \hat{p}(\boldsymbol\mu_{t+1})
\end{align}

\subsection{Update}
At the update step the predicted distribution~(\ref{eq:prediction}) of the latent variables for the next timestamp is corrected with the new data $\mathbf{y}_{t+1}$
\begin{align}
p(&\mathbf{x}_{t+1}, \boldsymbol\omega_{t+1}, \boldsymbol\gamma_{t+1}, \boldsymbol\mu_{t+1} | \mathbf{y}_{1:t+1})\nonumber \\
= & \dfrac{1}{Z} p(\mathbf{y}_{t+1} | \mathbf{x}_{t+1}, \boldsymbol\omega_{t+1}, \boldsymbol\gamma_{t+1}, \boldsymbol\mu_{t+1}) \nonumber\\
&\times p(\mathbf{x}_{t+1}, \boldsymbol\omega_{t+1}, \boldsymbol\gamma_{t+1}, \boldsymbol\mu_{t+1} | \mathbf{y}_{1:t}) \nonumber\\
= & \dfrac{1}{Z} p(\mathbf{y}_{t+1} | \mathbf{x}_{t+1}) p(\mathbf{x}_{t+1} | \boldsymbol\omega_{t+1}) p(\boldsymbol\omega_{t+1} | \boldsymbol\gamma_{t+1}) \nonumber\\
&\times p(\boldsymbol\gamma_{t+1} | \boldsymbol\mu_{t+1}) \hat{p}(\boldsymbol\mu_{t+1}),
\end{align}
where $Z$ is the normalisation constant.

Since components of the vectors $\mathbf{x}_{t+1}$ and $\boldsymbol\omega_{t+1}$ are conditionally independent, the terms $p(\mathbf{x}_{t+1} | \boldsymbol\omega_{t+1})$ and $p(\boldsymbol\omega_{t+1} | \boldsymbol\gamma_{t+1})$ are further factorised:
\begin{align}
p(&\mathbf{x}_{t+1}, \boldsymbol\omega_{t+1}, \boldsymbol\gamma_{t+1}, \boldsymbol\mu_{t+1} | \mathbf{y}_{1:t+1})\nonumber \\
= & \dfrac{1}{Z} p(\mathbf{y}_{t+1} | \mathbf{x}_{t+1}) \left[\prod_{i = 1}^N p(x_{i t+1} | \omega_{i t+1}) p(\omega_{i t+1} | \gamma_{i t+1})\right] \nonumber\\
\label{eq:update}
&\times p(\boldsymbol\gamma_{t+1} | \boldsymbol\mu_{t+1}) \hat{p}(\boldsymbol\mu_{t+1}),
\end{align}

The resulting formula for update~(\ref{eq:update}) is the same as the posterior distribution~(\ref{eq:P}) with the only exception in the term related to $\boldsymbol\mu_t$. The approximation of this posterior is proposed in Section \ref{sec:model}. The algorithm is only required to be adjusted for the new factor $\hat{p}(\boldsymbol\mu_{t+1})$.

The factor $\hat{p}(\boldsymbol\mu_{t+1})$ is a Gaussian distribution, i.e. it is from the exponential family already and it only depends on a single latent variable, therefore this factor should not be updated in the EP iterations. The information from this factor will be passed through the general approximating distribution $q$ to the other factors.

In the EP algorithm used for inference of the updated distribution~(\ref{eq:update}) the distribution for $\boldsymbol\mu_t$ is approximated with the Gaussian distribution for any $t$. Therefore, the identity~(\ref{eq:mu_int}) is true for any $t$ and the whole procedure can be applied for all timestamps.

\subsection{Minibatch filtering}
The developed Bayesian filtering procedure can be easily extended to the case of inferring minibatches for timestamps $[t+1:t+M]$, where $M$ is the size of a minibatch:
\begin{equation}
p(\mathbf{x}_{t+1:t+M}, \boldsymbol\omega_{t+1:t+M}, \boldsymbol\gamma_{t+1:t+M}, \boldsymbol\mu_{t+1:t+M} | \mathbf{y}_{1:t+M})
\end{equation}
rather than for the next timestamp $t+1$ only as in~(\ref{eq:update}).

Indeed, due to conditional independence marginalisation~(\ref{eq:predict_int}) also comes down to integral~(\ref{eq:mu_int}) similar to~(\ref{eq:predict_int_mu}). And the update step can also be performed by the EP algorithm with the only difference that it should be applied for $M$ timestamps rather than one.

%

\section{Experiments}
\label{sec:experiments}
This section presents validation and evaluation results for the proposed algorithms. The performance of these two-level GP algorithms is compared with:
\begin{itemize}
	\item the spatio-temporal spike and slab model with a one-level GP prior and its modification with common precision approximation~\cite{andersen2015bayesian};
	\item a popular alternating direction method of multipliers~(ADMM) method~\cite{boyd2011distributed}, which is a convex optimisation method used here for the lasso problem~\cite{Tibshirani96};
	\item a spatio-temporal sparse Bayesian learning~(STSBL) algorithm~\cite{zhang2014spatiotemporal}.
\end{itemize}

For quantitative comparison, the following measures are used:
\begin{itemize}
\item $\text{NMSE} (\text{normalised mean square error})= \dfrac{\|\mathbf{X} - \widehat{\mathbf{X}}\|^2_{F}}{\|\mathbf{X}\|^2_{F}}$, where $\mathbf{X}$ is the true signal, $\widehat{\mathbf{X}}$ is the estimate, computed as the mean of the approximated posterior distribution, $\|\cdot\|_{F}$ is the Frobenius norm of a matrix;
\item $\text{F-measure}~\cite{murphy2012machine} = 2\dfrac{\text{precision}\cdot\text{recall}}{\text{precision} + \text{recall}}$ between non-zero elements of the true signal $\mathbf{X}$ and non-zero elements of the estimate $\widehat{\mathbf{X}}$.
\end{itemize}
The NMSE shows the normalised error of signal reconstruction, with 0 corresponding to an ideal match. The F-measure shows how well slab locations are restored. An F-measure equal to 1 means that the true and estimated signals coincide, whilst 0 corresponds to lack of similarity between them. Arguably, for the sparse regression problem, the NMSE is less meaningful than the F-measure~\cite{xin2016maximal}.

Both two-level and one-level GP algorithms are iterated until convergence, which is measured by difference in the estimate of the signal $\widehat{\mathbf{X}}$ at the current and previous iterations.

\subsection{Synthetic data}

\begin{figure}[!t]
\centering
\subfloat[Data]{
	\includegraphics[width=0.2\textwidth]{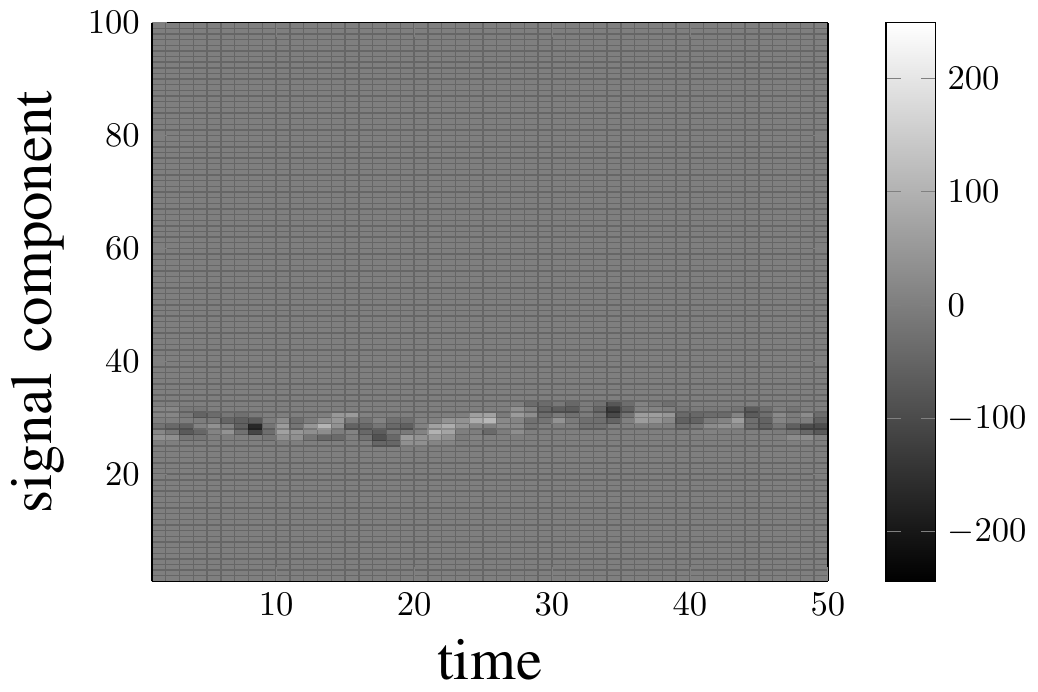}
	\label{pic:data_1}
}
\hfil
\subfloat[Data]{
	\includegraphics[width=0.2\textwidth]{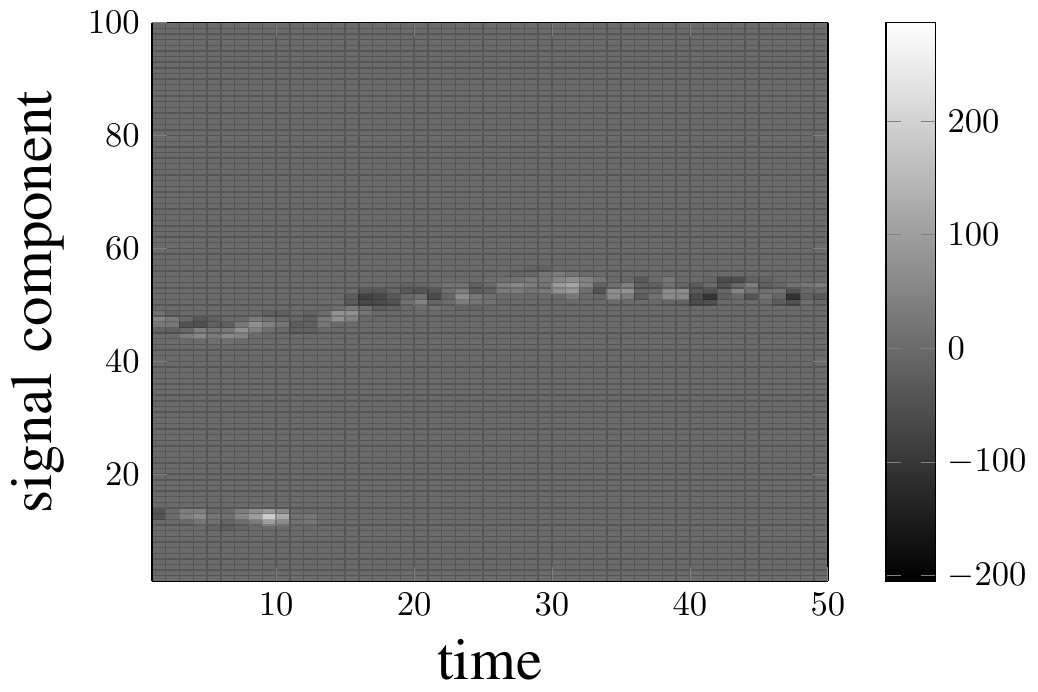}
	\label{pic:data_2}
}
\caption{Examples of the true signal $\mathbf{X}$ for the synthetic data. In each example two groups of slabs generated at $t=1$ evolve in time until $t=50$. }
\label{pic:synthetic_data}
\end{figure}

In this experiment, the algorithm performance is studied on synthetic data with known true values of signal $\mathbf{X}$ and slab locations $\boldsymbol{\Omega}$. The synthetic data represents the signals that have slowly evolving in time groups of non-zero elements. To create a spatio-temporal structure of slabs at the first timestamp $t=1$ two groups of slab locations are generated with Poisson-distributed sizes for the signal $\mathbf{x}_t$ of dimensionality $N = 100$. Then, from $t=2$ to $t=T=50$, these groups randomly evolve: each border of each group can go up, down, or stay at the same location with such probabilities that in average the sparsity level remains~$95\%$. In such way, locations of the slab groups are generated. The values of non-zero elements of the signal are then drawn from the distribution~$\mathcal{N}(0, 10^4)$. This procedure is repeated $10$ times to generate $10$ data samples. The examples of generated $\mathbf{X}$ are shown in \figurename~\ref{pic:synthetic_data}.

The elements of the design matrix $\mathbf{A}$ are generated as independent and identically distributed (iid) samples from the standard Gaussian. For each of the data samples, observations $\mathbf{Y} = \mathbf{A}\mathbf{X}$ of different length $K$ are generated. The value $K / N$ is referred as an undersampling ratio. It changes from $10\%$ to $55\%$.

\begin{figure}[!t]
\centering
\subfloat[F-measure]{
	\includegraphics[width=0.35\textwidth]{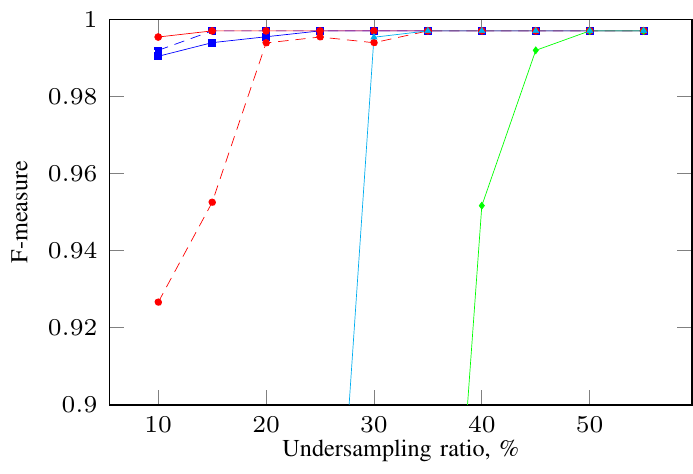}
	\label{pic:results_comparison_f_measure}
}
\hfil
\subfloat[NMSE]{
	\includegraphics[width=0.35\textwidth]{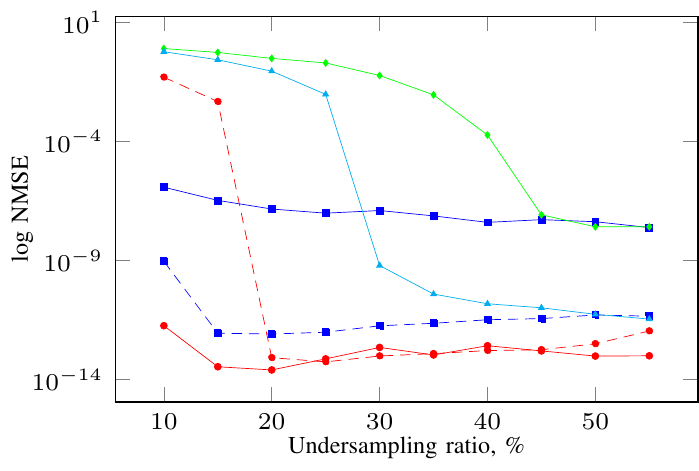}
	\label{pic:results_comparison_nmse}
}
\hfil
\subfloat[Time]{
	\includegraphics[width=0.35\textwidth]{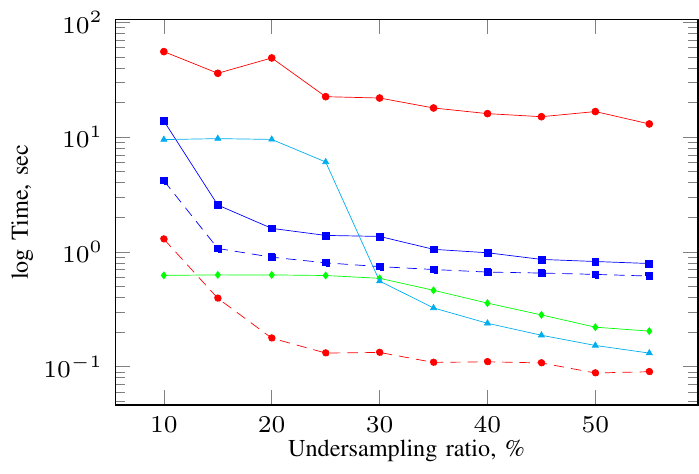}
	\label{pic:results_comparison_time}
}
\hfil
\subfloat{
	\includegraphics[width=0.35\textwidth]{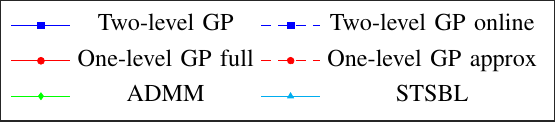}
	\label{pic:results_comparison_legend}
}

\caption[Caption for synthetic results figure]{Performance of the algorithms on the synthetic data. Note that the NMSE plots have logarithmic scale of y-axis. As the convergence criteria is $\dfrac{||\widehat{\mathbf{X}}^{\text{new}}-\widehat{\mathbf{X}}^{\text{old}}||_{\infty}}{||\widehat{\mathbf{X}}^{\text{old}}||_{\infty}} < 10^{-3}$, values below $10^{-3}$ are less significant. The proposed algorithms referred as two-level GP and two-level GP online outperform others in the $10-20\%$ interval, where the number of observations is the lowest.}
\label{pic:synthetic_results}
\end{figure}

The algorithms are evaluated in terms of average F-measure, NMSE and time\footnote{Time is evaluated with 4.2GHz Intel Core i7 CPU and 16GB RAM.} (\figurename~\ref{pic:synthetic_results}) on this data. On the interval between $10\%$ and $20\%$ of the undersampling ratio both inference methods for the two-level GP model and full EP inference for the one-level GP model show competitive results in terms of the accuracy metrics while outperforming the other methods. On the interval between $20\%$ and $30\%$ of the undersampling ratio the inference methods for one- and two-level GP models are already able to perfectly reconstruct the sparse signal while both ADMM and STSBL show less accurate results. STSBL achieves the perfect reconstruction starting from the undersampling ratio $30\%$ and ADMM achieves these results starting from the undersampling ratio $50\%$.

In the proposed EP algorithm for the two-level GP model (Section \ref{sec:model}), the complexity of each iteration is $\mathcal{O}(N^3T)$, as matrices of size $N\times N$ are inverted for each timestamp to compute cavity distributions for the factors $u$ and $r$.
In the proposed online inference algorithm (Section~\ref{sec:online_inference}), first the offline version is trained on size $T_\text{init}$. Then, when new data of size $M$ is available, the previous results are used as prior and the complexity of update is $\mathcal{O}(N^3M)$, while in the offline version it is $\mathcal{O}(N^3(T_\text{init}+M))$.

On average, the proposed two-level GP algorithm requires similar to the full one-level GP algorithm number of iterations for convergence: approximately $30$ iterations on the interval between $10\%$ and $20\%$ of the undersampling ratio, $15$ iterations on the interval between $20\%$ and $30\%$, and less than $10$ iterations for the higher undersampling ratios. The approximate inference algorithm for the one-level GP model takes slightly more iterations to converge.

In the one-level GP algorithm \cite{andersen2015bayesian} the complexity of one iteration is $\mathcal{O}(N^3T^3)$. This is related to inversion of full spatio-temporal covariance matrix. It is addressed with low rank and common precision approximations \cite{andersen2015bayesian}, which reduce both the computational complexity and the quality of the results. The $K$-rank approximation, where $K$ is a parameter of the algorithm, reduces the computational complexity to $\mathcal{O}(N^2KT)$ and the common precision approximation reduces it to $\mathcal{O}(N^2T + T^2N)$.

In terms of the computational time the full EP inference for the one-level GP model is the slowest method. The approximated inference for the one-level GP model significantly improve its performance in terms of the computational time while also cause loss in accuracy. The ADMM method shows similar results to the approximated one-level GP model in terms of the computational time, but has even bigger loss in terms of both accuracy measures. The STSBL takes slightly more time for the lower values of the undersampling ratio, which helps it to achieve better results than the ADMM method in terms of the accuracy measures. The proposed offline and online inference methods for the two-level GP method demonstrate a satisfactory trade-off between computational time and accuracy. They obtain competitive results in terms of accuracy measures as the full EP inference for the one-level GP model while require significantly less computational time. In terms of computational time the proposed method demonstrates competitive results with the STSBL method.

The proposed online inference method for the two-level GP model allows to save computational time while preserving the accuracy of the recovered signal. Note that the developed inference methods for the two-level GP model outperform competitors in the lowest undersampling ratio interval, i.e. they require less measurements to get the same quality as other algorithms.

\subsection{Real data: moving object detection in video}

The considered methods for sparse regression are compared on the problem of object detection in video sequences. The Convoy dataset~\cite{Warnell2014} is used where a background frame is subtracted from each video frame. As moving objects take only part of a frame the considered signal of the subtracted video frames is sparse. Moreover, objects are represented as clusters of pixels, which evolve in time. Therefore, the background subtraction application fully satisfies the proposed spatio-temporal structured model assumptions.

The frames with subtracted background are resized to~$32 \times 32$ pixels and reshaped as vectors~$\mathbf{x}_t \in \mathbb{R}^N$, $N = 1024$. The number of frames in the dataset is~$T = 260$. The sparse observations are obtained as $\mathbf{Y} = \mathbf{A} \mathbf{X}$, where $\mathbf{A} \in \mathbb{R}^{K\times N}$ is the matrix with iid Gaussian elements. $10$ different random design matrices $\mathbf{A}$ are used to generate $10$ data samples. The number of observations $K$ is chosen such that the undersampling ratio $K / N$ changes from $10\%$ to $55\%$.
This procedure corresponds to compressive sensing observations~\cite{Cevher2008}.

For this problem the full EP inference for the one-level GP model is infeasible due to its memory requirements, therefore only the common precision approximated inference for the one-level GP model is considered.

\begin{figure}[!t]
\centering
\subfloat[F-measure]{
	\includegraphics[width=0.35\textwidth]{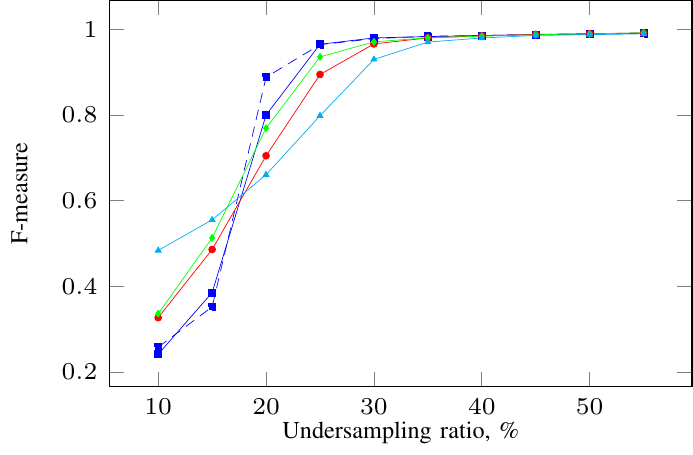}
	\label{pic:results_comparison_f_measure_convoy}
}
\hfil
\subfloat[NMSE]{
	\includegraphics[width=0.35\textwidth]{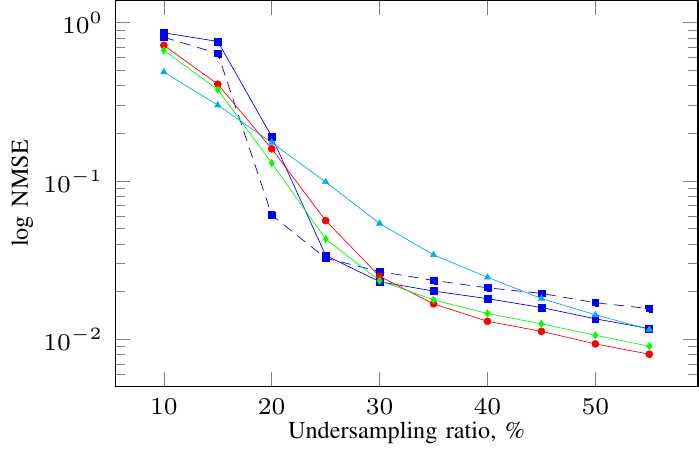}
	\label{pic:results_comparison_nmse_convoy}
}
\hfil
\subfloat{
	\includegraphics[width=0.35\textwidth]{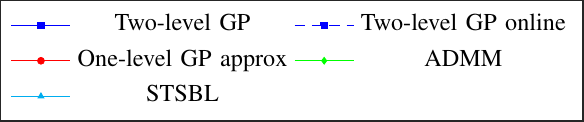}
	\label{pic:results_comparison_legend_convoy}
}

\caption[Caption for convoy results figure]{Performance of the algorithms on the Convoy data. The proposed algorithms referred as two-level GP and two-level GP online outperform the others in the $20-30\%$ interval. On the interval $10-15\%$ all methods cannot reconstruct the true signal. The NMSE plot shows that the proposed algorithms underperform the competitors for the values higher than $30\%$, but the visual difference in performance becomes insignificant that is demonstrated in \figurename~\ref{pic:reconstruction}.}
\label{pic:convoy_results}
\end{figure}

The average F-measure and NMSE obtained by all the algorithms on the Convoy data are presented in~\figurename~\ref{pic:convoy_results}. The proposed algorithm shows the best results for the undersampling ratio $20 - 30\%$. For larger values of the undersampling ratio all the algorithms provide close almost ideal results of reconstruction.

\begin{figure*}[!t]
\centering
\subfloat[Original frame]{\includegraphics[width=0.15\textwidth]{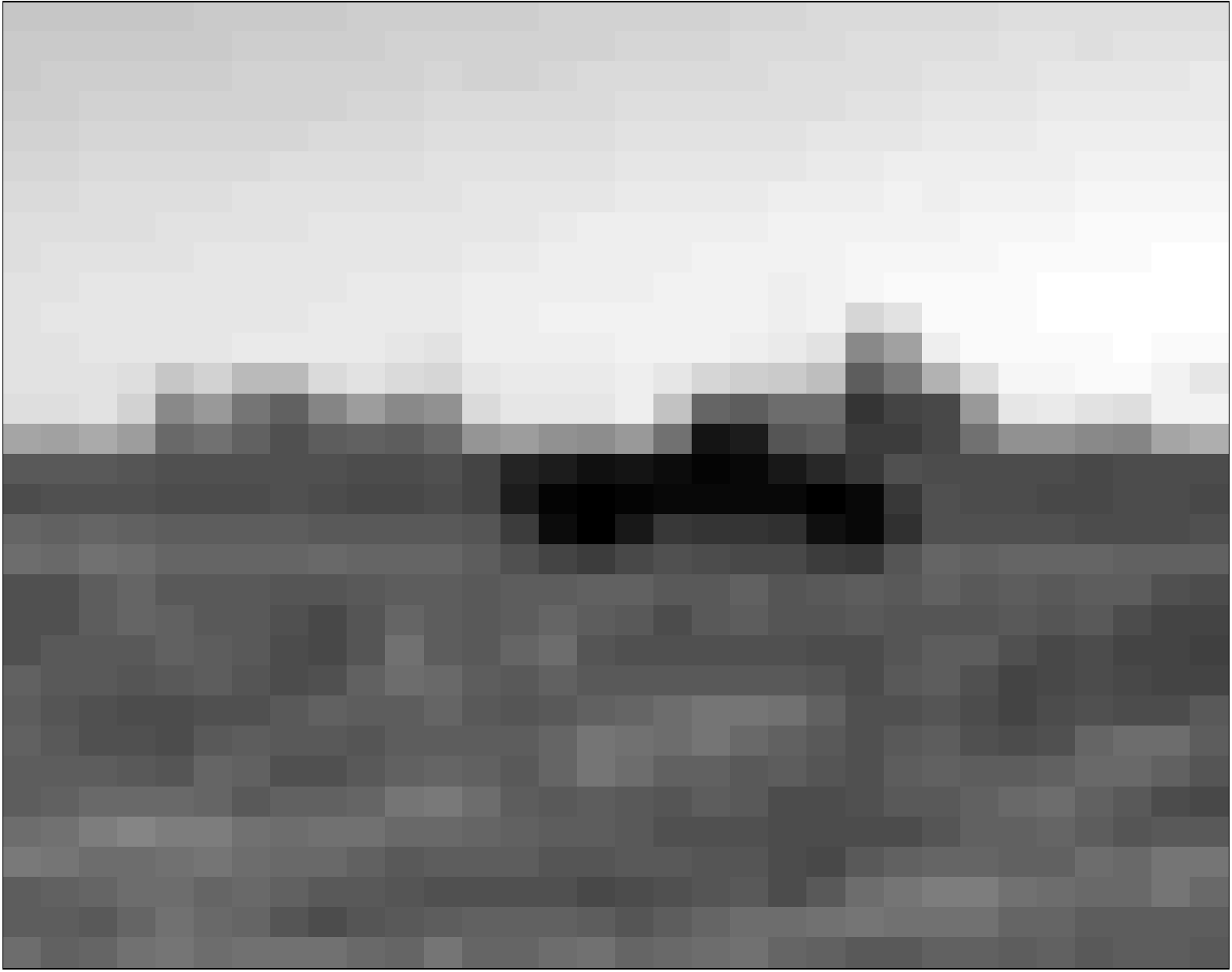}
\label{subfig:original}}
\hfil
\subfloat[Two-level GP $10\%$]{\includegraphics[width=0.15\textwidth]{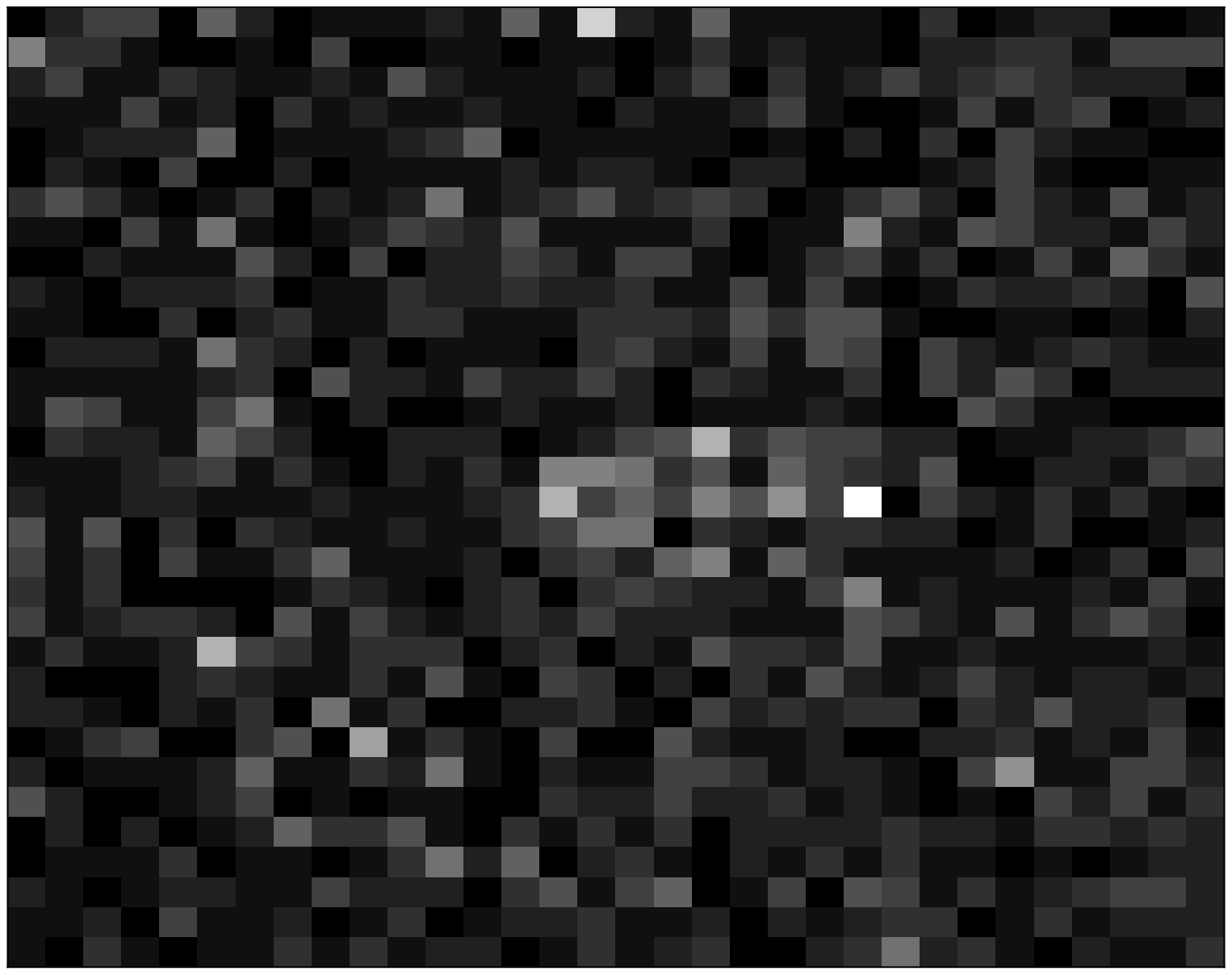}
\label{subfig:our_car_10}}
\hfil
\subfloat[One-level GP $10\%$]{\includegraphics[width=0.15\textwidth]{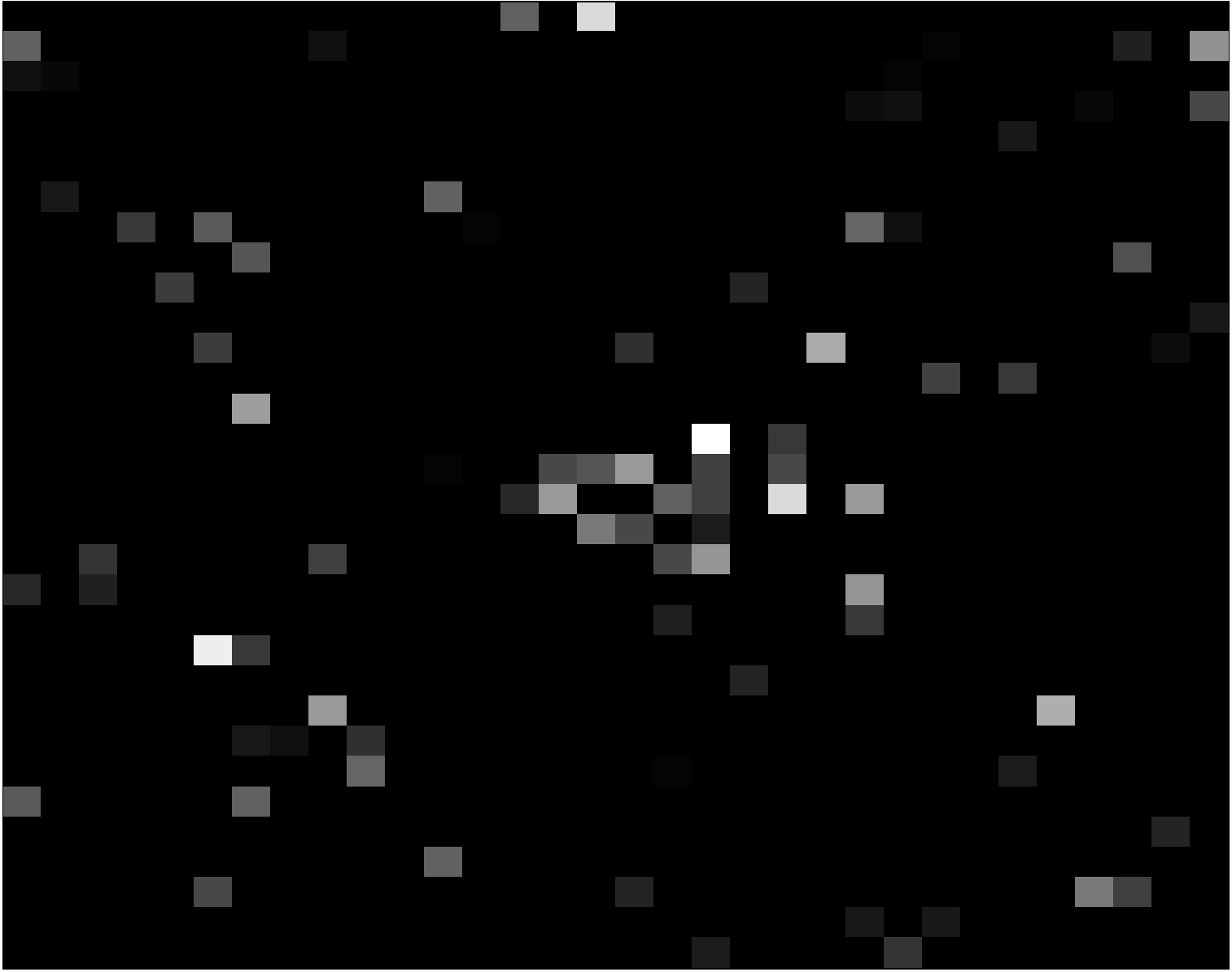}
\label{subfig:andersen_car_10}}
\hfil
\subfloat[ADMM $10\%$]{\includegraphics[width=0.15\textwidth]{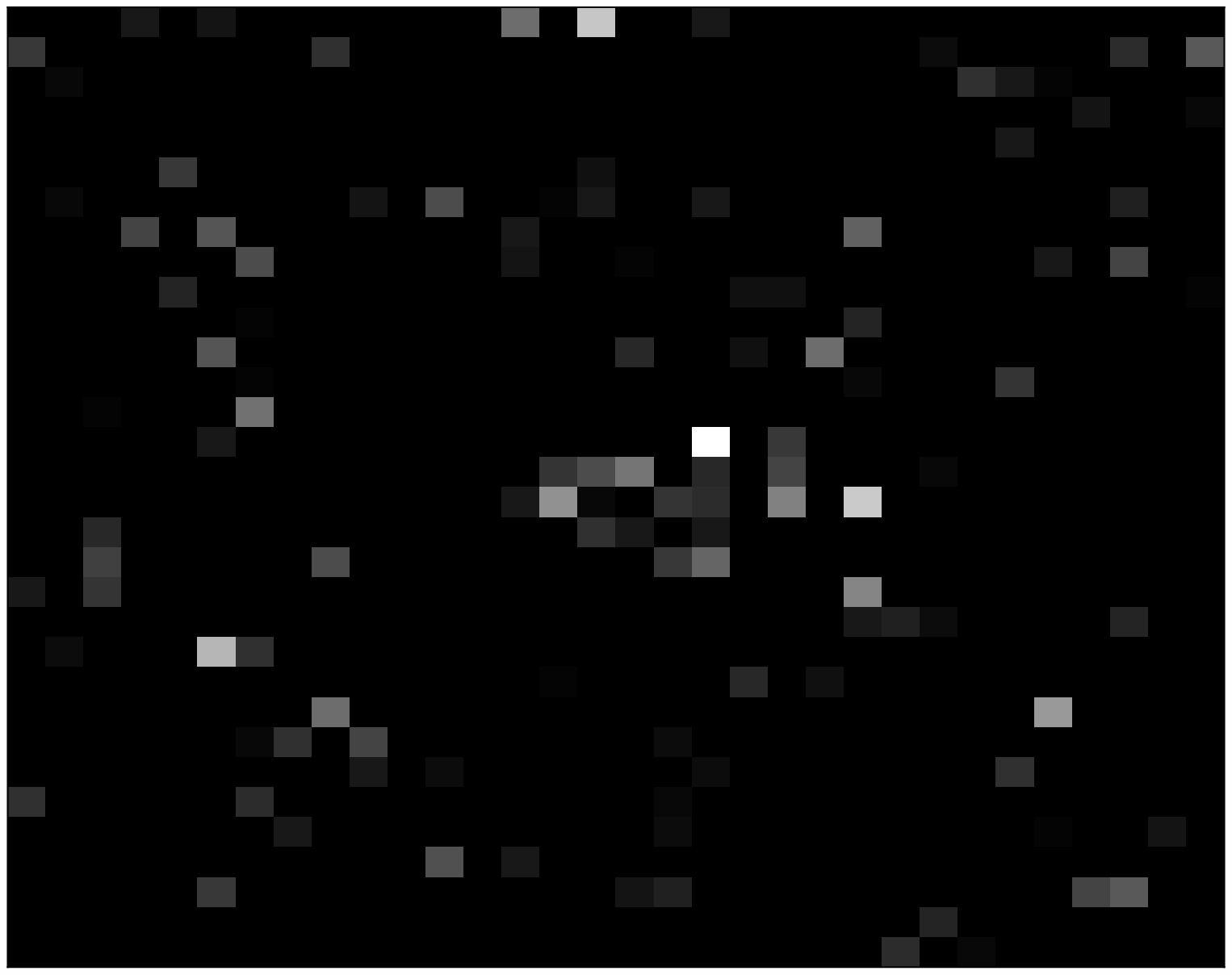}
\label{subfig:lasso_car_10}}
\hfil
\subfloat[STSBL $10\%$]{\includegraphics[width=0.15\textwidth]{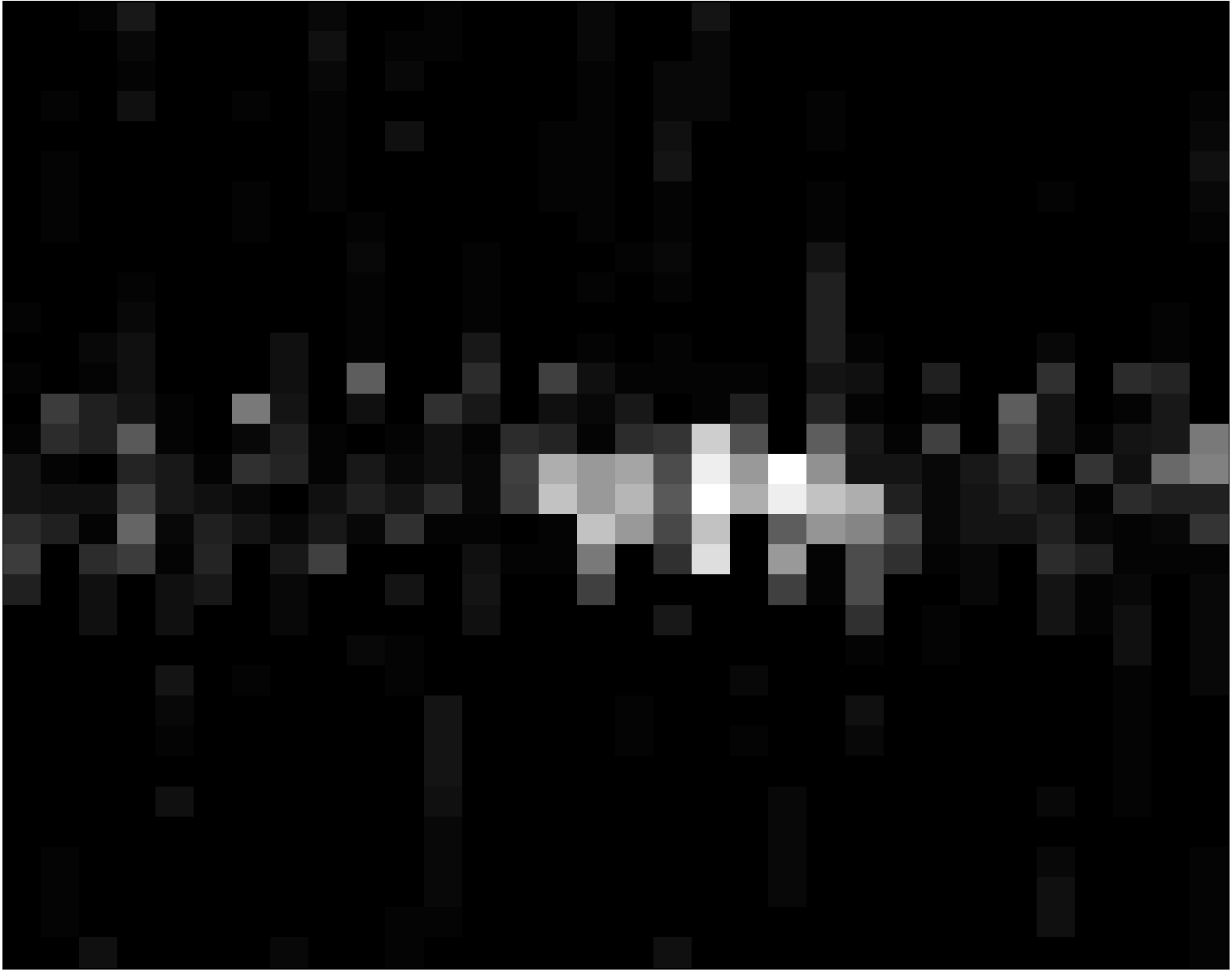}
\label{subfig:stsbl_car_10}}
\\
\subfloat[Background frame]{\includegraphics[width=0.15\textwidth]{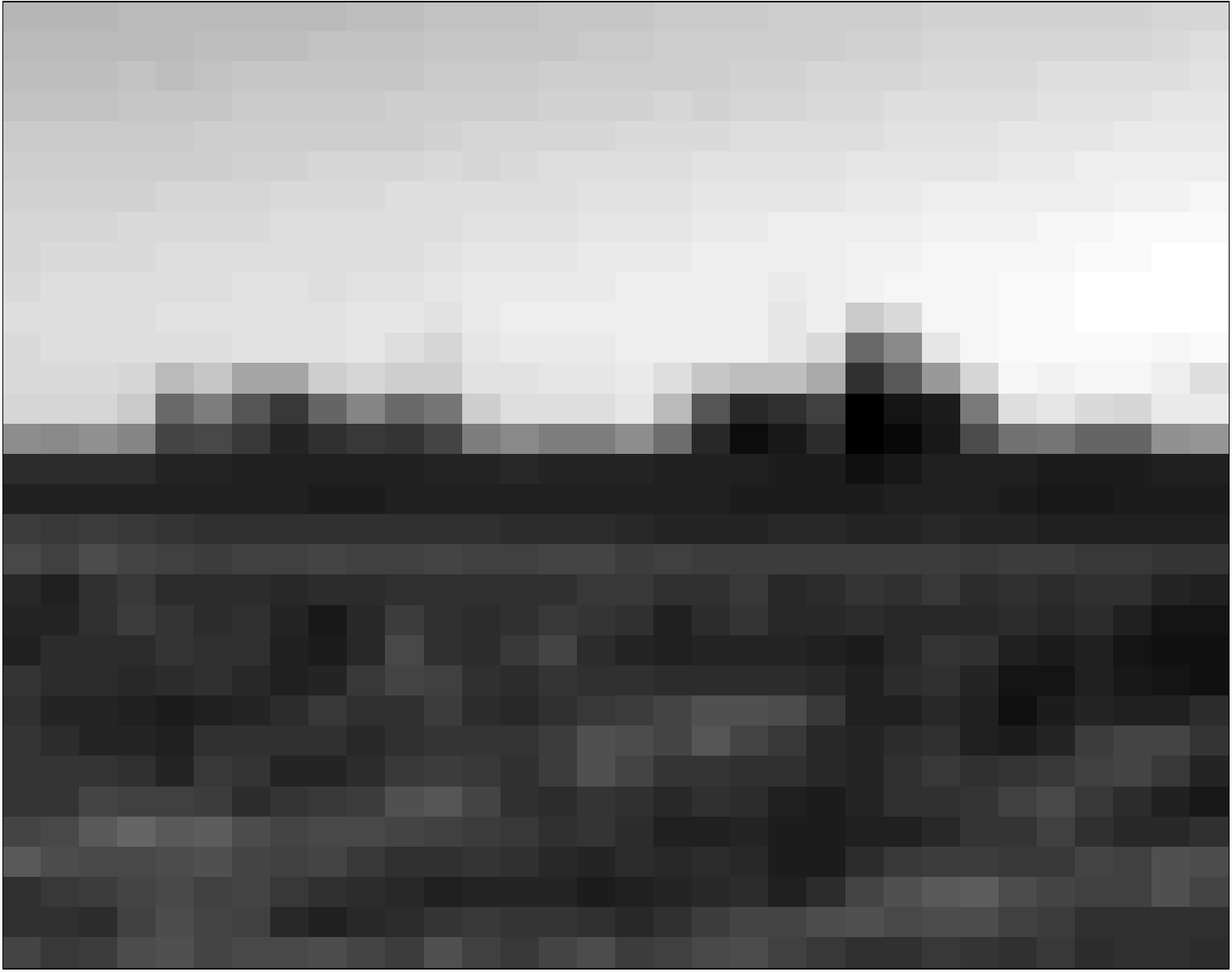}
\label{subfig:background}}
\hfil
\subfloat[Two-level GP $20\%$]{\includegraphics[width=0.15\textwidth]{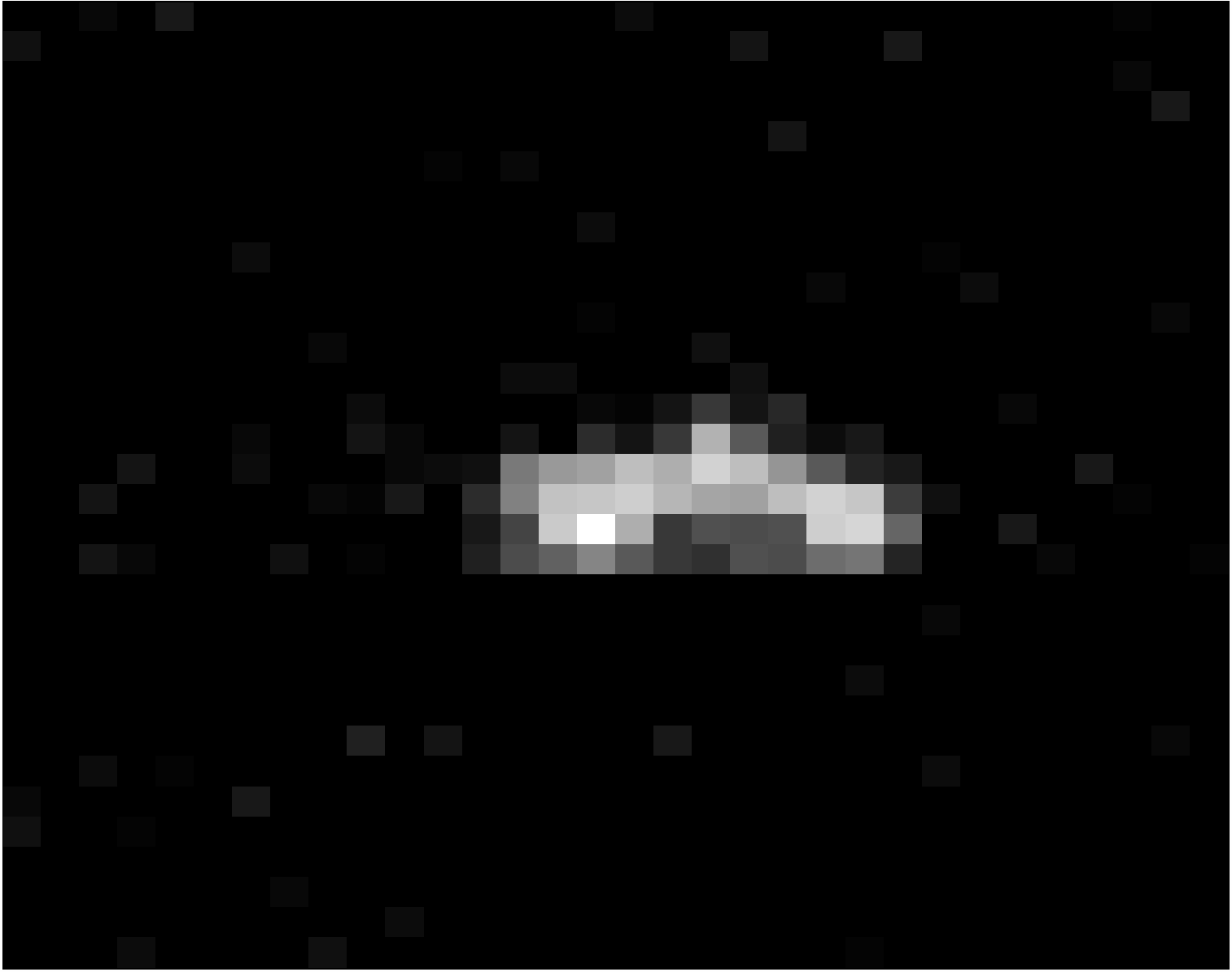}
\label{subfig:our_car_20}}
\hfil
\subfloat[One-level GP $20\%$]{\includegraphics[width=0.15\textwidth]{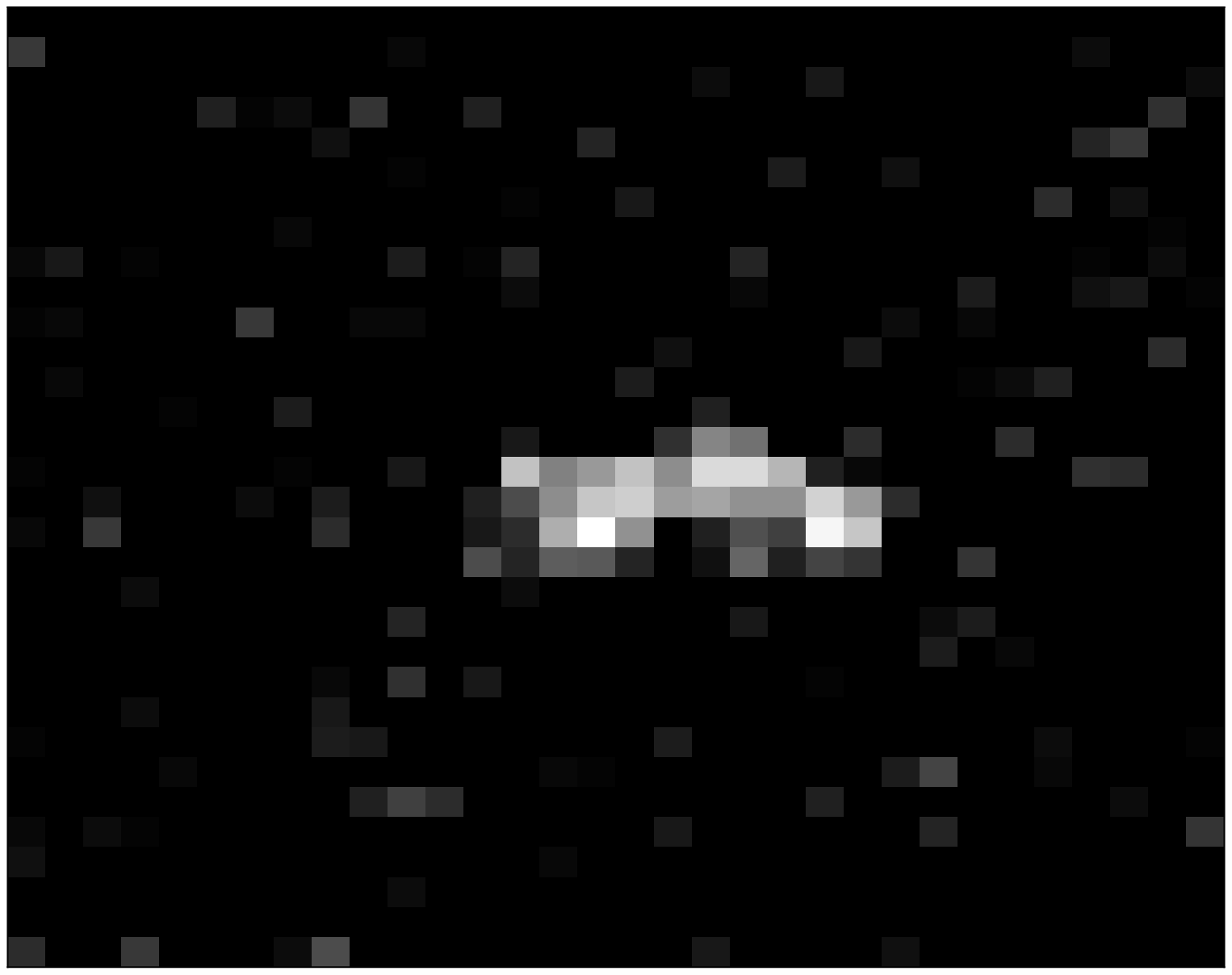}
\label{subfig:andersen_car_20}}
\hfil
\subfloat[ADMM $20\%$]{\includegraphics[width=0.15\textwidth]{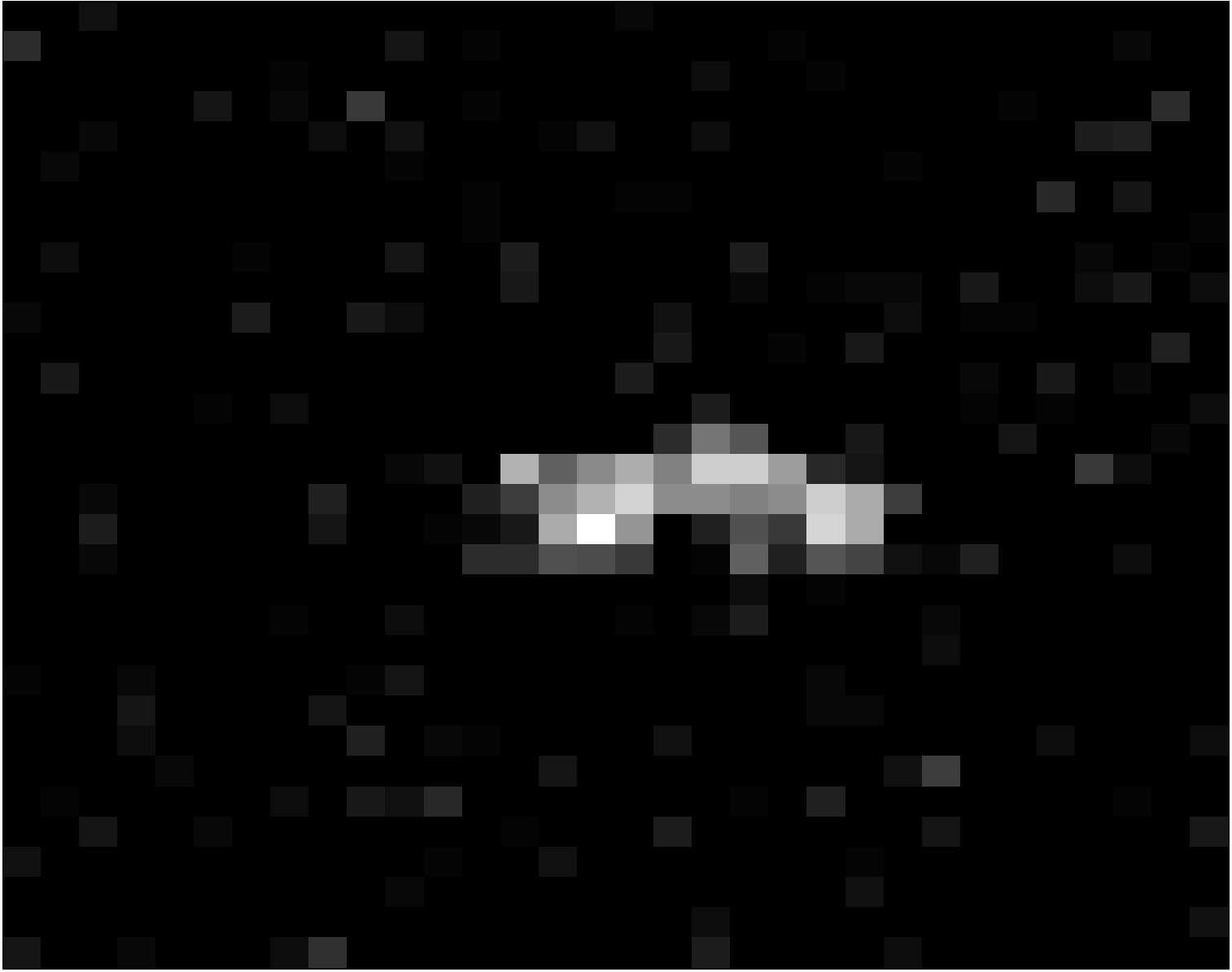}
\label{subfig:lasso_car_20}}
\hfil
\subfloat[STSBL $20\%$]{\includegraphics[width=0.15\textwidth]{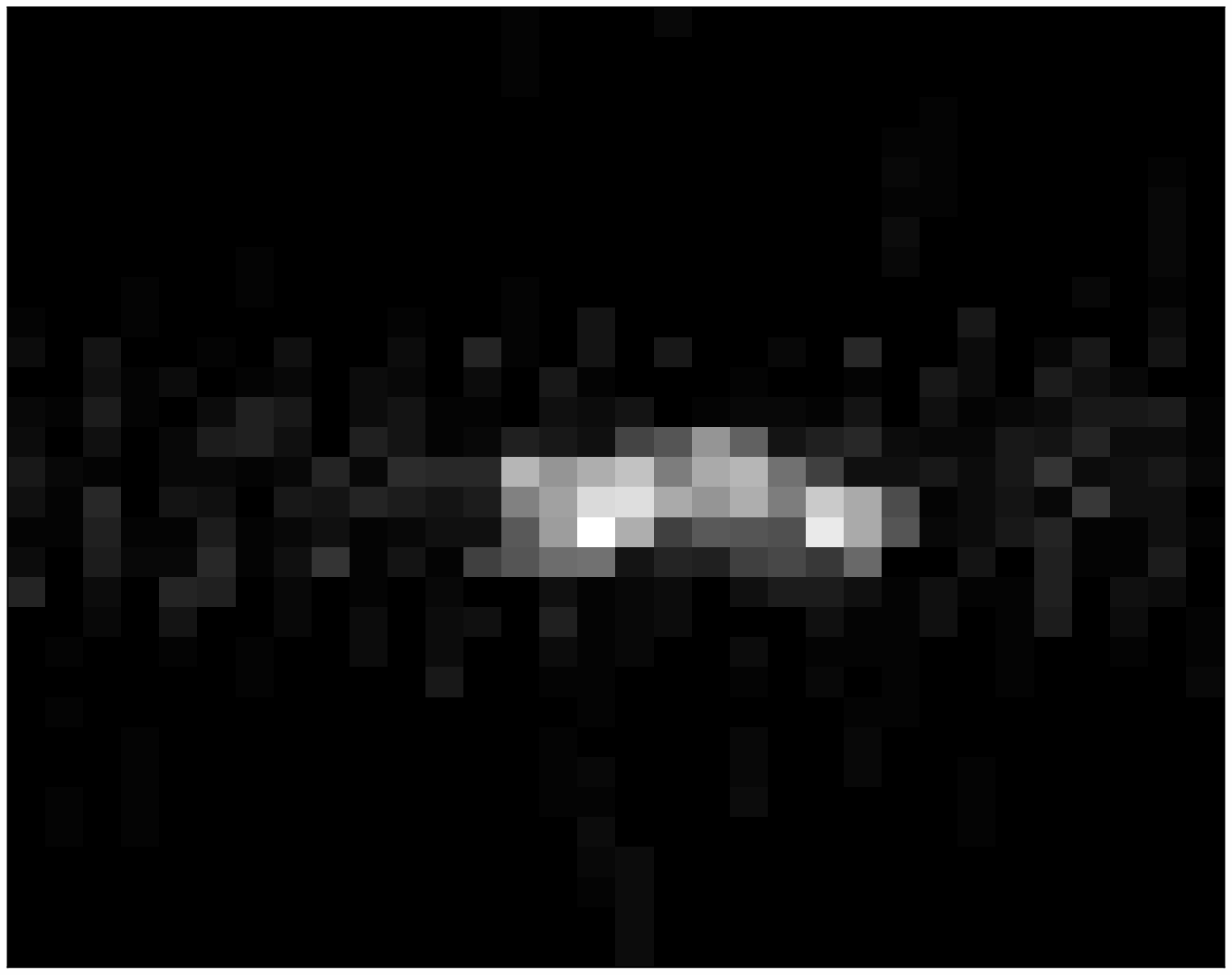}
\label{subfig:stsbl_car_20}}
\\
\subfloat[Reference object detection]{\includegraphics[width=0.15\textwidth]{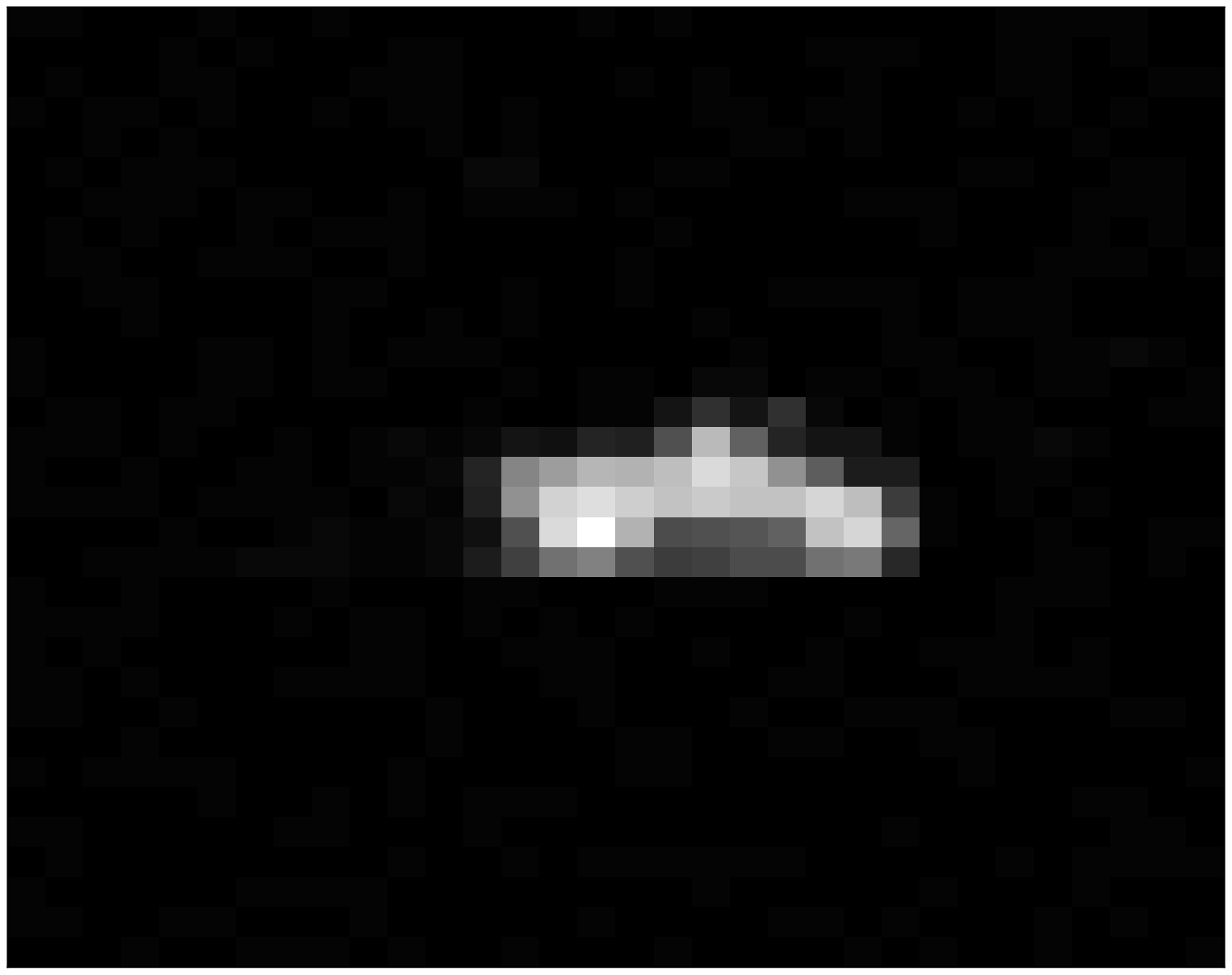}
\label{subfig:true_subtracted}}
\hfil
\subfloat[Two-level GP $40\%$]{\includegraphics[width=0.15\textwidth]{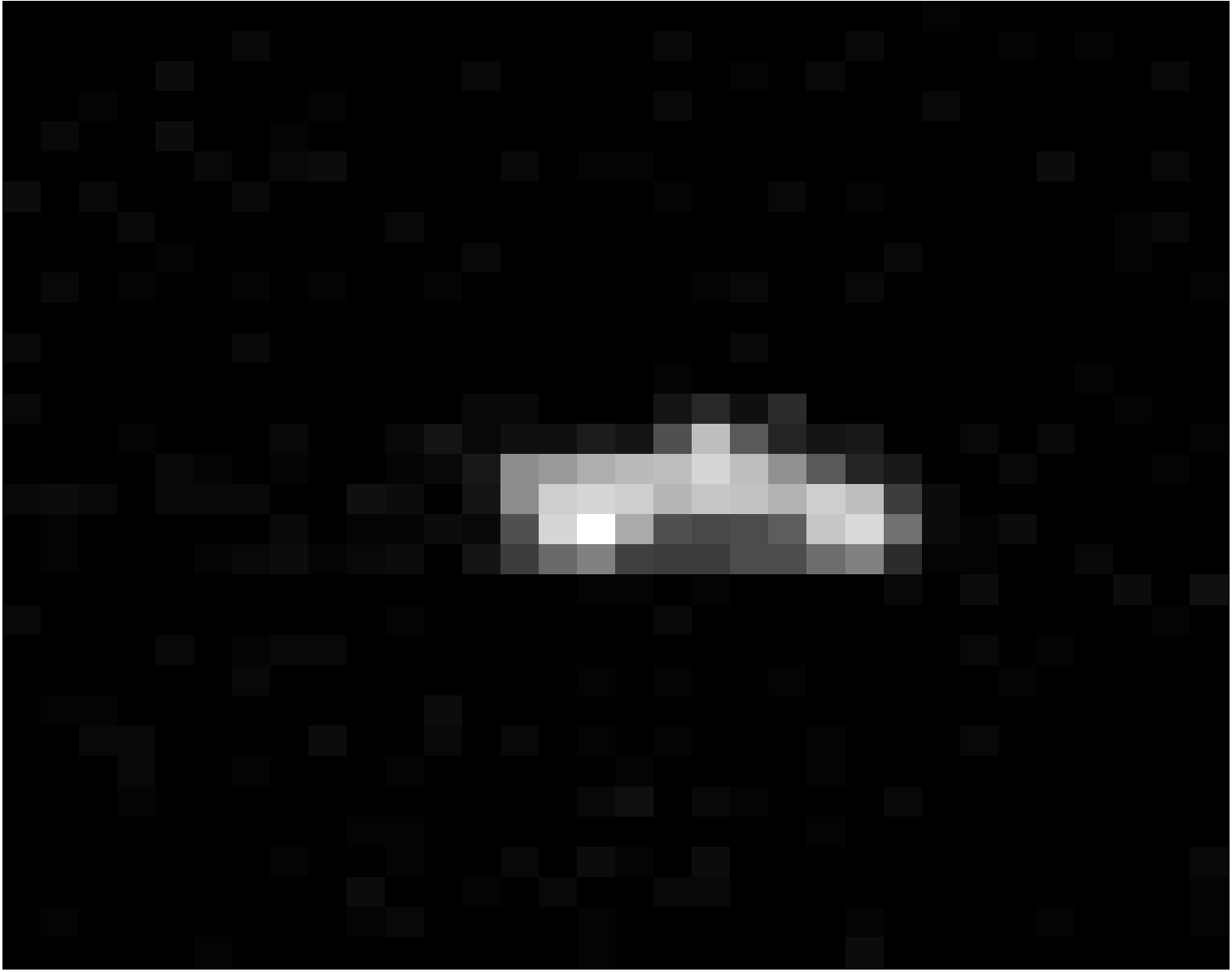}
\label{subfig:our_car_40}}
\hfil
\subfloat[One-level GP $40\%$]{\includegraphics[width=0.15\textwidth]{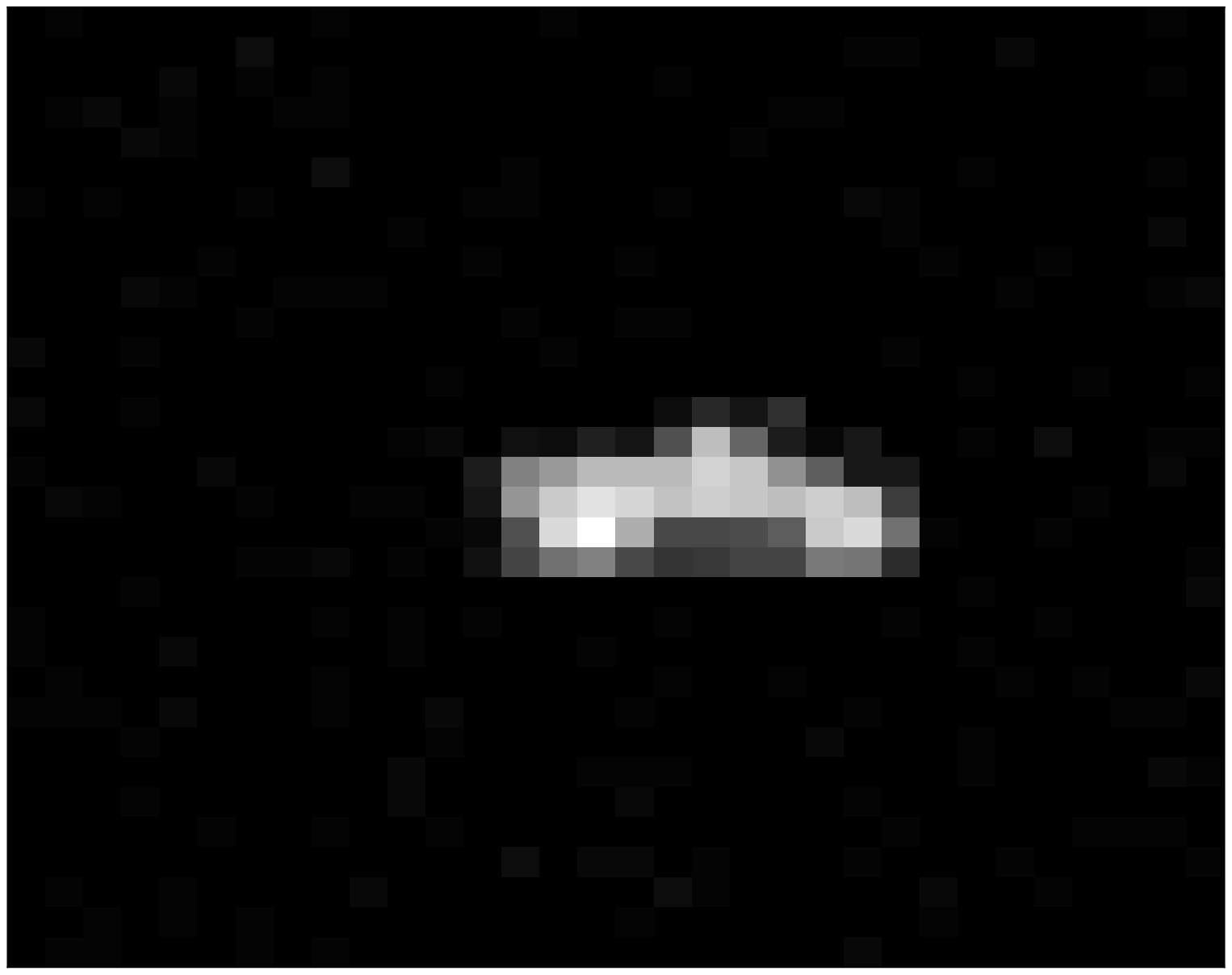}
\label{subfig:andersen_car_40}}
\hfil
\subfloat[ADMM $40\%$]{\includegraphics[width=0.15\textwidth]{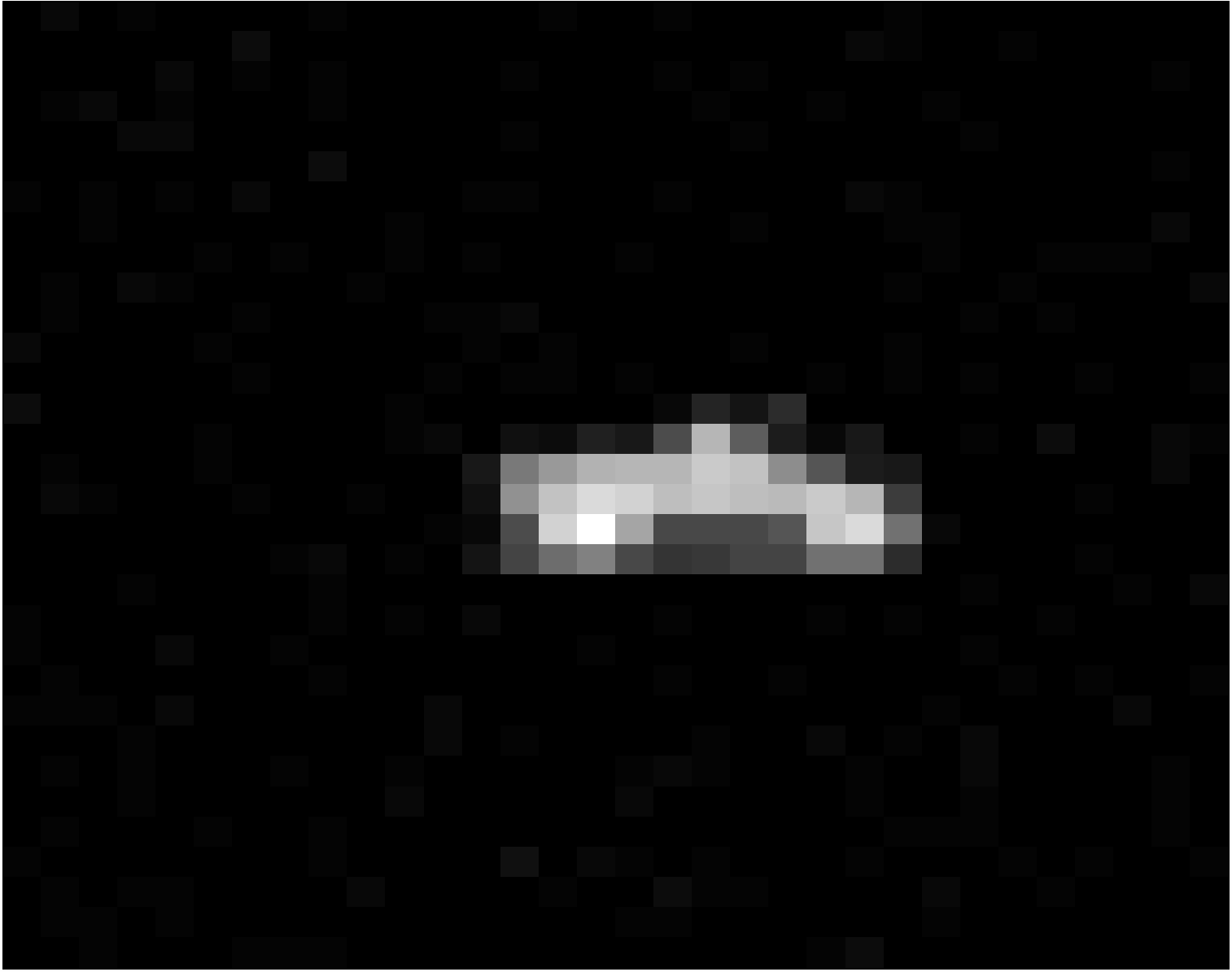}
\label{subfig:lasso_car_40}}
\hfil
\subfloat[STSBL $40\%$]{\includegraphics[width=0.15\textwidth]{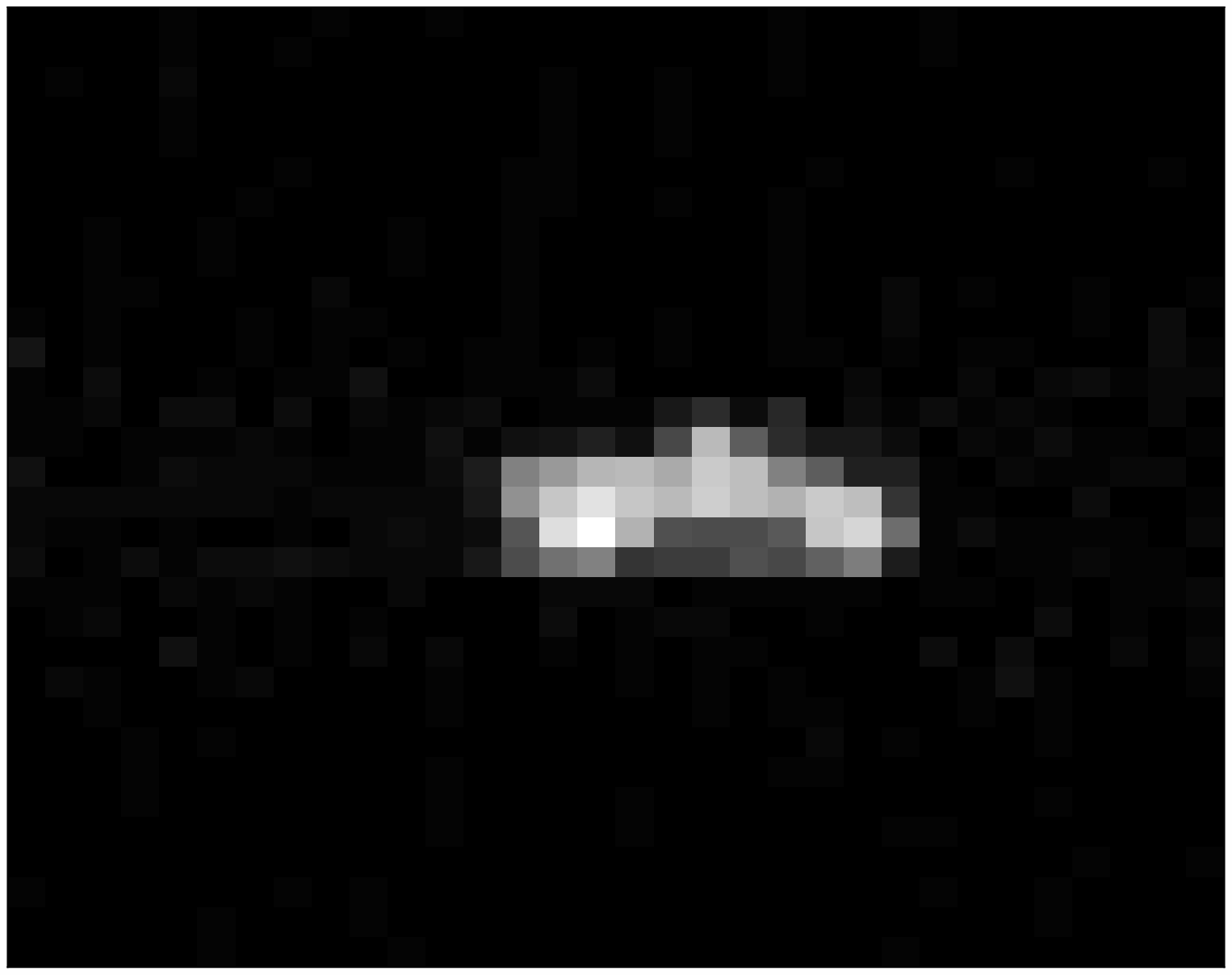}
\label{subfig:stsbl_car_40}}

\caption{Sample frame with reconstruction results from sparse observations for the Convoy data. \protect\subref{subfig:original}, \protect\subref{subfig:background}: the original and static background non-compressed frames; \protect\subref{subfig:true_subtracted}: object detection results based on non-compressed frame difference (static background frame is subtracted from the original frame); \protect\subref{subfig:our_car_10}, \protect\subref{subfig:our_car_20}, \protect\subref{subfig:our_car_40}: reconstruction of compressed object detection results based on the proposed online two-level GP method; \protect\subref{subfig:andersen_car_10}, \protect\subref{subfig:andersen_car_20}, \protect\subref{subfig:andersen_car_40}: reconstruction of the compressed object detection results based on the one-level GP method; \protect\subref{subfig:lasso_car_10}, \protect\subref{subfig:lasso_car_20}, \protect\subref{subfig:lasso_car_40}: reconstruction of the compressed object detection results based on the ADMM method; \protect\subref{subfig:stsbl_car_10}, \protect\subref{subfig:stsbl_car_20}, \protect\subref{subfig:stsbl_car_40}: reconstruction of the compressed object detection results based on the STSBL method. \protect\subref{subfig:our_car_10}, \protect\subref{subfig:andersen_car_10}, \protect\subref{subfig:lasso_car_10}, and \protect\subref{subfig:stsbl_car_10} show the results for the undersampling rate $10\%$, where all the algorithms fail to reconstruct the true signal. \protect\subref{subfig:our_car_20}, \protect\subref{subfig:andersen_car_20}, \protect\subref{subfig:lasso_car_20}, and \protect\subref{subfig:stsbl_car_20} show the reconstruction for the undersampling rate $20\%$, where the difference in performance between the algorithms is visible. While for the undersampling rate $40\%$ (\protect\subref{subfig:our_car_40}, \protect\subref{subfig:andersen_car_40}, \protect\subref{subfig:lasso_car_40}, and \protect\subref{subfig:stsbl_car_40}) reconstruction results are indistinguishable in quality.}
\label{pic:reconstruction}
\end{figure*}

\figurename~\ref{pic:reconstruction} presents the reconstructed sample frame from the Convoy data. For all the algorithms, the reconstruction results are provided for the undersampling ratio $10\%$, where the proposed algorithms slightly underperform the competitors in terms of the quality metrics, for the undersampling ratio $20\%$, where the proposed algorithm outperforms the competitors both in terms of NMSE and the F-measure, and for the undersampling ratio $40\%$, where the proposed algorithms show a little higher NMSE. It is clearly seen that for the undersampling ratio $10\%$ the difference in the quality metrics is insignificant since none of the methods is able to reconstruct the signal. The STSBL represents an exceptional example but still the frame reconstructed by this method contains considerable amount of noise. For the undersampling ratio $20\%$ the proposed method provides the clear reconstructed frame in contrast to the reconstructed frames by all the competitors that are more noisy. Meanwhile, for the undersampling ratio $40\%$ the difference between reconstruction results by all four algorithms is not remarkable.

Note that similar to the synthetic data experiment the proposed algorithms obtain the best results for the lowest undersampling ratio values where the reconstruction is reasonable, i.e. they require a less number of observations.

\subsection{Real data: EEG source localisation}
The third experiment is devoted to the EEG source localisation problem.

The goal of the non-invasive EEG source localisation problem is to find 3D locations of dipoles such that their electromagnetic field coincides with the field measured by electrodes on the human head cortex. This is important, for example, for localisation of active areas in human-brain interfaces and treatment of neurological disorders~\cite{jatoi2014survey, baillet2001electromagnetic}. This problem is ill-posed in sense that there exist an infinite number of possible active areas inside the brain that could produce the same field on the head cortex. To regularise the problem, we use the idea that slab locations are distributed in space and temporally evolve, similar to~\cite{baillet1997bayesian}. Similar idea applies to the MEG source localisation \cite{solin2016regularizing}.

Using the earlier introduced notation, the EEG source localisation problem is stated as
\begin{equation}
\mathbf{y}_t=\mathbf{A}\mathbf{x}_t + \boldsymbol\varepsilon_t, \quad \forall t\in [1, \ldots, T],
\end{equation}
where $\mathbf{y}_t\in\mathbb{R}^{K}$ is the vector containing observations of potential differences taken from $K = 69$ electrodes placed on a human head cortex, $\mathbf{A}\in\mathbb{R}^{K\times N}$ is the lead field matrix corresponding to $N/3 = 272$ voxels, $\mathbf{x}_t\in\mathbb{R}^{N}$ is the signal, that is the current density of dipole activation.

Here $\mathbf{x}_t$ represents the dipole moments corresponding to the grid locations:
\begin{equation}
\mathbf{x}_{t} =
\left[
x_{1x},
x_{1y},
x_{1z},
x_{2x},
x_{2y},
x_{2z},
\ldots,
x_{\frac{N}{3}z}
\right]^{\top}.
\end{equation}
For each grid voxel $i$ inside the brain with location coordinates $loc(i) = (x_i, y_i, z_i)$ the corresponding dipole moments $(x_{ix}, x_{iy}, x_{iz})$ along the 3D axis are considered.

We employ the following covariance function that promotes close values for collinear dipole moments corresponding to close grid positions
\begin{equation}
\mathbf{K}(i,j) = \alpha_{\mathbf{K}} \exp\left(-\frac{d(i,j)^2}{2\ell_{\mathbf{K}}^2}\right), \quad \mathbf{K} \in \{\boldsymbol\Sigma_0, \mathbf{W}\},
\end{equation}
where the distance is computed as
\begin{equation}
d(i,j) =
\begin{cases}
0, \text{if axis for dipole moments $i$, $j$ are different} \\
|| loc(i) - loc(j) ||^2_2, \text{otherwise}.
\end{cases}
\end{equation}

Hyperparameters are selected so that the sampled potential differences have the similar behaviour as the provided data.

The data and lead field matrix for the experiments is processed with EEGLAB~\cite{delorme2004eeglab}. We use the data provided in EEGLAB for the source localisation problem with annotated events.

\begin{figure}[!t]
\centering
\subfloat[Located dipole moments 1 ms after the event]{\includegraphics[width=0.45\columnwidth]{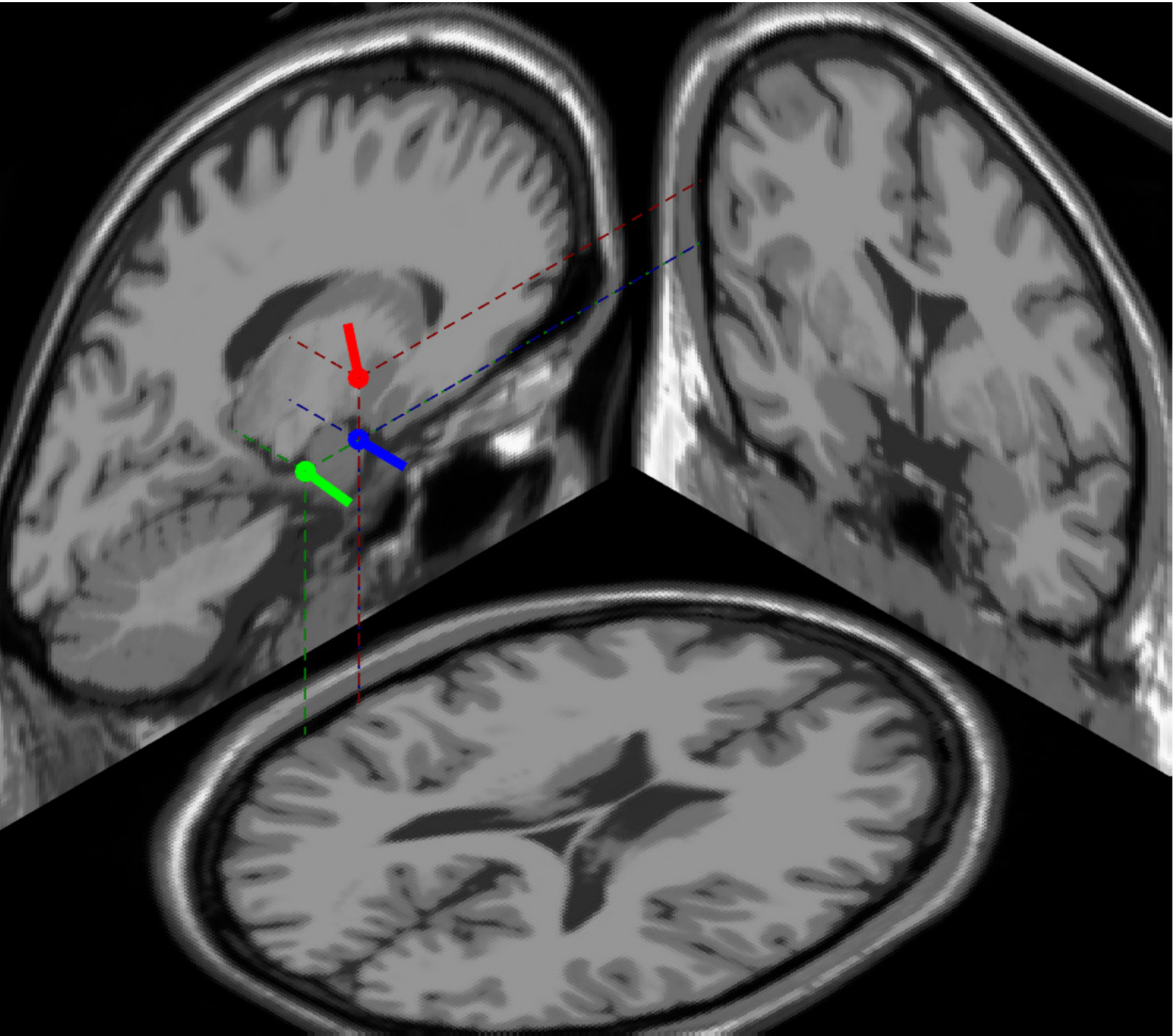}
\label{subfig:eeg_time_1}}
\hfil
\subfloat[Located dipole moments 170 ms after the event]{\includegraphics[width=0.45\columnwidth]{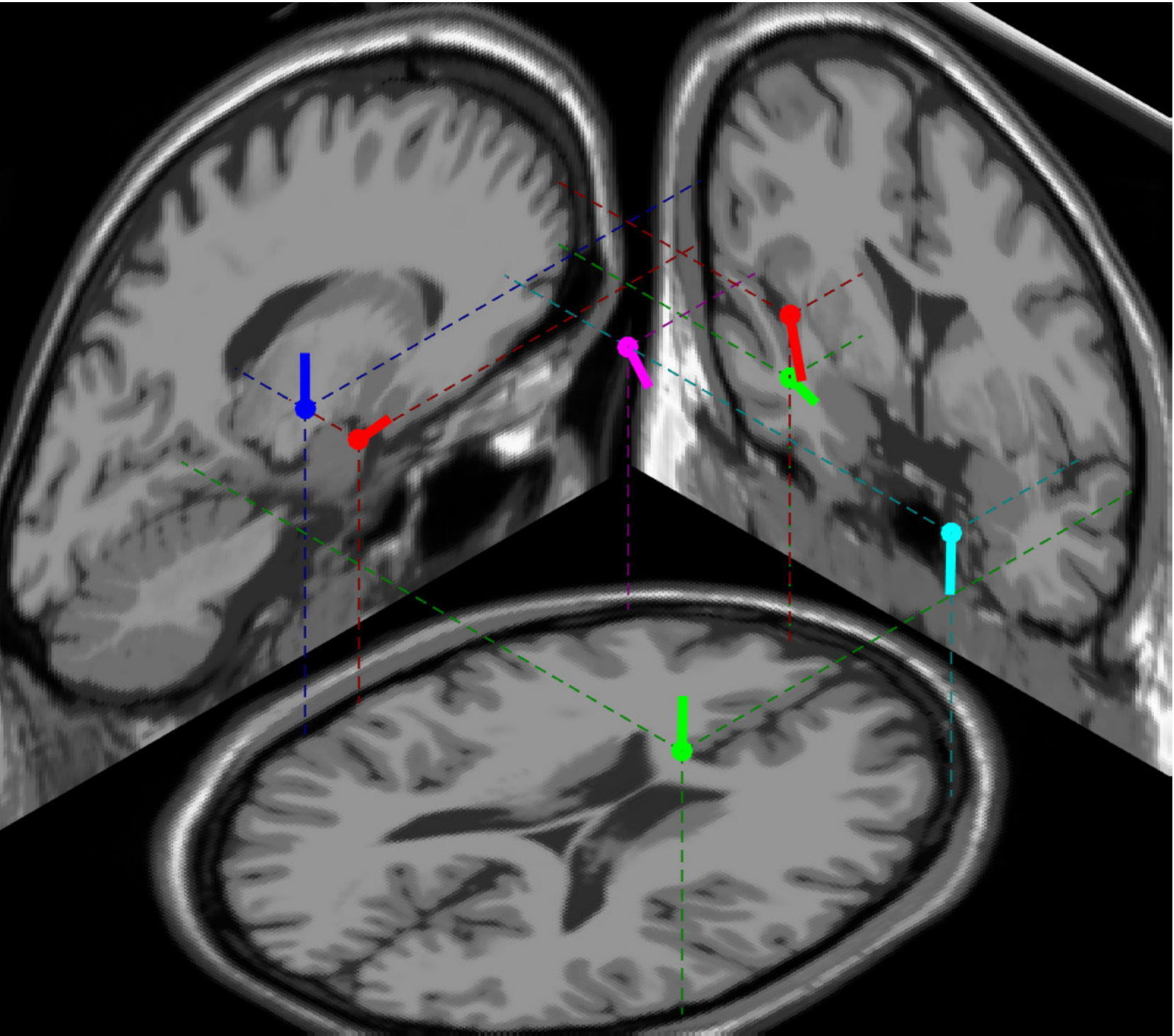}
\label{subfig:eeg_time_170}}
\caption{Located dipoles by the proposed offline two-level GP method for the EEG source localisation problem. There is no brain response immediately after the event and \protect\subref{subfig:eeg_time_1} demonstrates reconstructed brain active area that remains active during the whole period and it is not related to the event. While \protect\subref{subfig:eeg_time_170} shows the reconstructed active area when the brain response to the event is detected.}
\label{fig:eeg_located_moments}
\end{figure}

\begin{figure}[!t]
\centering
\subfloat[Measured EEG]{\includegraphics[width=0.5\columnwidth]{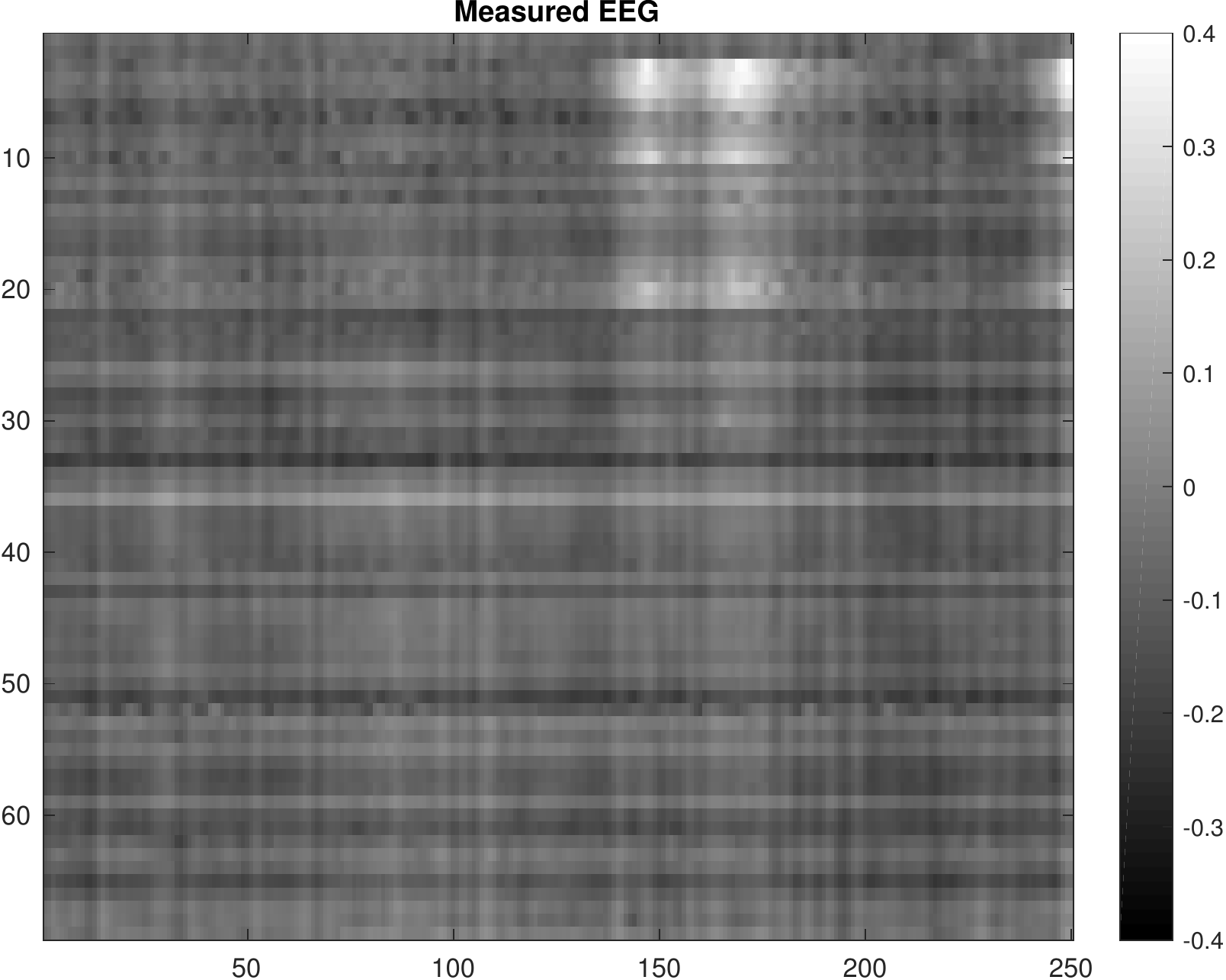}}
\hfil
\subfloat[Reconstructed EEG ($\mathbf{A}\hat{\mathbf{X}}$)]
{\includegraphics[width=0.5\columnwidth]{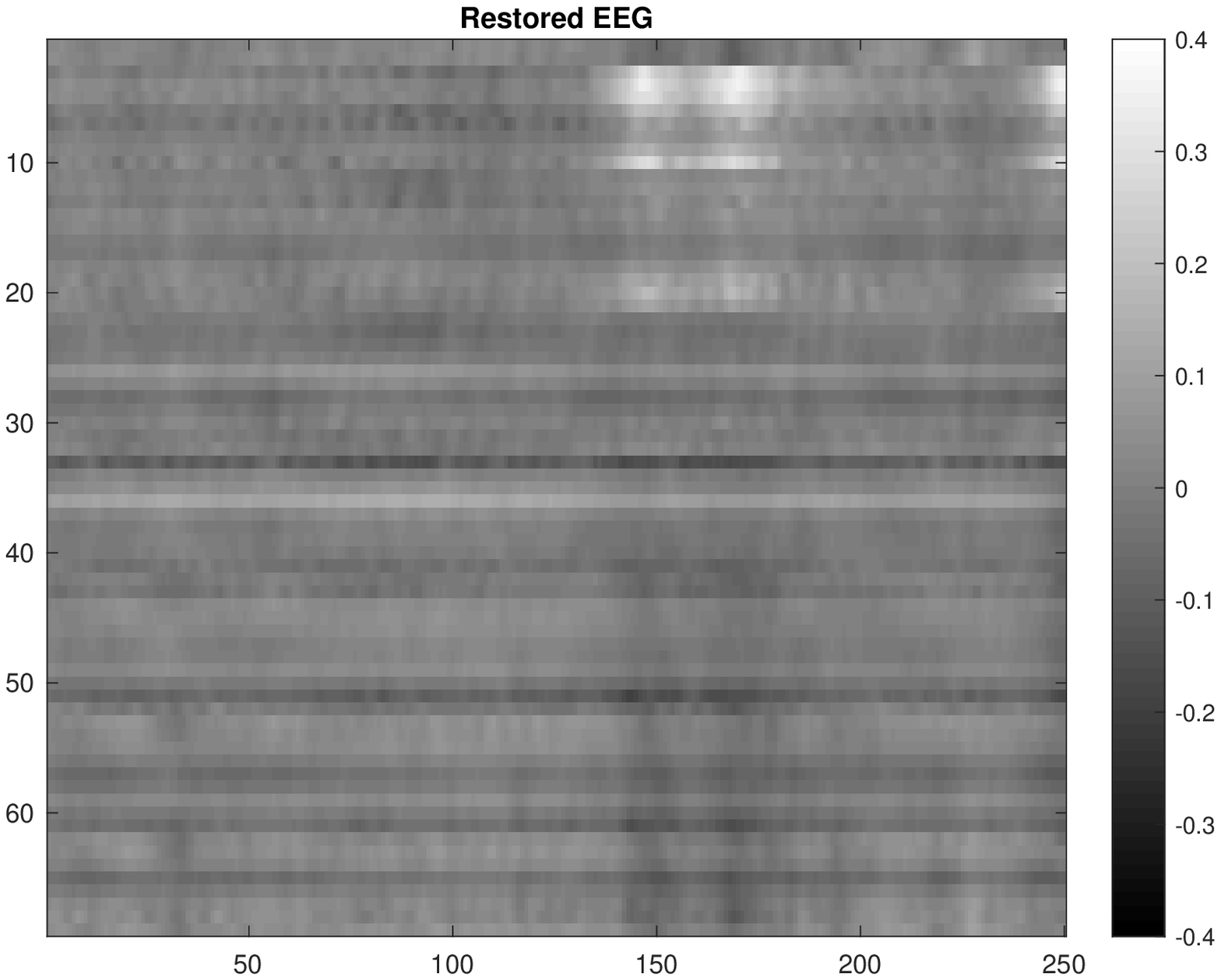}}
\caption{Reconstruction by the proposed offline two-level GP method of the EEG signal. As the true active dipole areas are not known, reconstruction quality is based on the observations $\mathbf{Y}$. Reconstructed EEG has lower magnitude, potentially because noise has been taken into account.}
\label{fig:eeg_restoration}
\end{figure}

Figure~\ref{fig:eeg_located_moments} presents located dipoles by the proposed method for the fourth event at two given time moments. The first time moment is taken right after the event happened and there is no response to it in the brain activity yet. The second time moment is chosen when the response is detected. Figure~\ref{fig:eeg_restoration} shows the comparison of measured and restored potential differences by the proposed algorithm.

\begin{figure}[!t]
\centering
\includegraphics{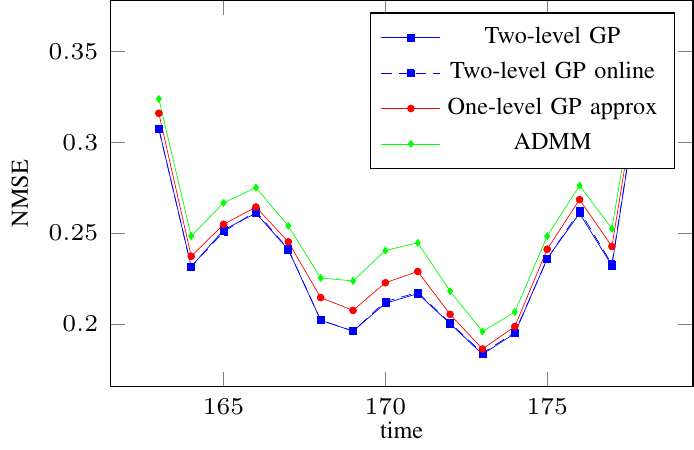}
\caption{Results for NMSE between $\mathbf{y}_t$ and $\mathbf{A}\widehat{\mathbf{x}}_t$ during the brain response time. The proposed algorithms referred as two-level GP and two-level GP online have the lowest NMSE among the others.}
\label{pic:eeg_data}
\end{figure}

The true signal $\mathbf{X}$ is unknown for the EEG source localisation problem, therefore, NMSE between the observations $\mathbf{y}_t$ and reconstructed $\mathbf{A}\widehat{\mathbf{x}}_t$  is used for the quantitative comparison in this experiment. The obtained results for all the algorithms around the time of the brain response are presented in~\figurename~\ref{pic:eeg_data}. The proposed two-level GP algorithms show the best results among the competitors. Both proposed offline and online inference methods demonstrate similar performance. Note that in this experiment the undersampling ratio is approximately $8\%$, which confirms that the proposed method is able to provide better results for lower values of the undersampling ratio.

\subsection{Parameters selection}
For the proposed algorithm and for the one-level GP the parameters $\eta$ and $\xi$ are grid optimised to make the comparison fair. The prior shape hyperparameters $\ell_{\Sigma}$, $\ell_{W}$, $\alpha_{\Sigma}$, $\alpha_{W}$ and variances $\sigma^2_x$ and $\sigma^2$ are specified so that sampled data has the same form as training data. ADMM and STSBL use the default values of parameters. The selected hyperparameter values for the proposed algorithm for all datasets are presented in Table~\ref{tab:our_hyperparameters}.
 \begin{table}[t]
  \caption{Two-level GP hyperparameters}
  \label{tab:our_hyperparameters}
  \centering
  \begin{tabular}{llll}
    \toprule
    Parameter & Synthetic & Convoy &  EEG\\
    \midrule
    $\sigma^2_x$ & $10^4$ & $160$ &  $4 * 10^5$\\
    $\sigma^2$ & $10^{-4}$ & $4$ &  $10^{-3}$\\
    $\eta$ & $0.999$ & $0.99$ &  $0.9$\\
    $\xi$ & $0.9999$ & $0.999$ &  $0.8$\\
    $\ell_W$ & $15$ & $15$ &  $22.17$\\
    $\ell_\Sigma$ & $10$ & $10$ &  $0.2217$\\
    $\alpha_W$ & $10$ & $10$ &  $10^{-2}$\\
    $\alpha_\Sigma$ & $10$ & $10$ &  $0.05$\\
    \bottomrule
  \end{tabular}
\end{table}

\section{Conclusions}
\label{sec:conclusions}
This paper proposes a new hierarchical Gaussian process model of spatio-temporal structure representation with complex temporal evolution in sparse Bayesian inference methods. This is achieved using the flexible hierarchical GP prior for the spike and slab model, where spatial and temporal structural dependencies are encoded by different levels of the prior. Offline and online methods are developed for posterior inference for this model.

We show that the introduced model can be applied to different areas such as compressive sensing and EEG source localisation. The results show the superiority of the proposed method in comparison with the non-hierarchical GP method, the alternating direction method of multipliers and the spatio-temporal sparse Bayesian learning method. The developed algorithms demonstrate better performance both in terms of signal value reconstruction and localisation of non-zero signal components: within the low amount of measurements range it achieves around 15\% improvement in terms of slab localisation quality.

\subsubsection*{Acknowledgments}
The authors would like to thank the support from the EC Seventh Framework Programme [FP7 2013-2017] TRAcking in compleX sensor systems (TRAX) Grant agreement no.: 607400.
%

\bibliographystyle{IEEEtran}
\bibliography{References}

\appendices
\section{Product and quotient rules}
\label{sec:app_rules}
EP updates are based on products and quotients of distributions. This section presents the product and quotient rules for Gaussian and Bernoulli distributions.

\subsection{Product of Gaussians}
A product of two Gaussian distributions is a unnormalised Gaussian distribution
\begin{equation*}
\mathcal{N}(\mathbf{x}; \mathbf{m}_1, \boldsymbol{\Sigma}_1) \mathcal{N}(\mathbf{x}; \mathbf{m}_2, \boldsymbol{\Sigma}_2) \propto \mathcal{N}(\mathbf{x}; \mathbf{m}, \boldsymbol{\Sigma}),
\end{equation*}
where
\begin{equation*}
\boldsymbol\Sigma^{-1} = \boldsymbol\Sigma_1^{-1} + \boldsymbol\Sigma_2^{-1}, \, \boldsymbol\Sigma^{-1} \mathbf{m} = \boldsymbol\Sigma_1^{-1} \mathbf{m}_1 + \boldsymbol\Sigma_2^{-1} \mathbf{m}_2
\end{equation*}

\subsection{Quotient of Gaussians}
A quotient of two Gaussian distributions is a unnormalised Gaussian distribution\footnote{Although quotient can lose positive semidefiniteness, we will still refer to it as a Gaussian distribution}
\begin{equation*}
\dfrac{\mathcal{N}(\mathbf{x}; \mathbf{m}_1, \boldsymbol{\Sigma}_1)}{\mathcal{N}(\mathbf{x}; \mathbf{m}_2, \boldsymbol{\Sigma}_2)} \propto \mathcal{N}(\mathbf{x}; \mathbf{m}, \boldsymbol{\Sigma}),
\end{equation*}
where
\begin{equation*}
\boldsymbol\Sigma^{-1} = \boldsymbol\Sigma_1^{-1} - \boldsymbol\Sigma_2^{-1}, \, \boldsymbol\Sigma^{-1} \mathbf{m} = \boldsymbol\Sigma_1^{-1} \mathbf{m}_1 - \boldsymbol\Sigma_2^{-1} \mathbf{m}_2
\end{equation*}

\subsection{Product of Bernoulli}
A product of two Bernoulli distributions is a unnormalised Bernoulli distribution
\begin{equation*}
\text{Ber}(x; \Phi(z_1)) \text{Ber}(x; \Phi(z_2)) \propto \text{Ber}(x; \Phi(t(z_1, z_2))),
\end{equation*}
where
\begin{equation*}
t(z_1, z_2) = \Phi^{-1}\left( \left[\dfrac{(1 - \Phi(z_1))(1 - \Phi(z_2))}{\Phi(z_1) \Phi(z_2)} + 1 \right]^{-1} \right)
\end{equation*}

\subsection{Quotient of Bernoulli}
A quotient of two Bernoulli distributions is a unnormalised Bernoulli distribution
\begin{equation*}
\dfrac{\text{Ber}(x; \Phi(z_1))}{\text{Ber}(x; \Phi(z_2))} \propto \text{Ber}(x; \Phi(d(z_1, z_2))),
\end{equation*}
where
\begin{equation*}
d(z_1, z_2) = \Phi^{-1}\left( \left[\dfrac{(1 - \Phi(z_1))\Phi(z_2)}{(1 - \Phi(z_2))\Phi(z_1)} + 1 \right]^{-1} \right)
\end{equation*}

\section{EP Update for factor $f_{it}$}
\label{sec:update_f}

\subsection{Cavity distribution}
The unnormalised cavity distribution $q^{\setminus q_{f_{it}}} \left(x_{it}, \omega_{it}\right) = \frac{q\left(x_{it}, \omega_{it}\right)}{q_{f_{it}}\left(x_{it}, \omega_{it}\right)}$ can be computed as
\begin{align*}
q^{\setminus q_{f_{it}}}&=\dfrac{\mathcal{N}(x_{it}; \mathbf{m}_t(i), \mathbf{V}_t(i, i))\text{Ber}(\omega_{it}; \Phi(z_{it}))}{\mathcal{N}(x_{it}; \mathbf{m}_{f_{t}}(i), \mathbf{V}_{f_{t}}(i, i))\text{Ber}(\omega_{it}; \Phi(z_{f_{it}}))} \\
&\propto\mathcal{N}(x_{it}; m_{it}^{\setminus f}, v_{it}^{\setminus f}) \text{Ber}(\omega_{it}; \Phi(z_{it}^{\setminus f})),
\end{align*}
where
\begin{flalign*}
(v_{it}^{\setminus f})^{-1} &= \mathbf{V}^{-1}_t(i, i) - \mathbf{V}^{-1}_{f_{t}}(i, i),\\
(v_{it}^{\setminus f})^{-1} m_{it}^{\setminus f} &= \mathbf{V}^{-1}_t(i, i) \mathbf{m}_t(i) - \mathbf{V}^{-1}_{f_{t}}(i, i) \mathbf{m}_{f_{t}}(i, i),\\
z_{it}^{\setminus f} &= z_{h_{it}}
\end{flalign*}

\subsection{Moments matching}
The moments of the tilted distribution $q^{\setminus q_{f_{it}}}f_{it}$ are
\begin{align*}
Z_{it} &= \Phi(z_{it}^{\setminus f}) \mathcal{N}(0;m_{it}^{\setminus f},v_{it}^{\setminus f})\nonumber\\
&+(1 - \Phi(z_{it}^{\setminus f})) \mathcal{N}(0;m_{it}^{\setminus f},v_{it}^{\setminus f}+ \sigma_x^2),\\
\mathbb{E} x_{it} &= \dfrac{1-\Phi(z_{it}^{\setminus f})}{Z_{it}}\mathcal{N}(0;m_{it}^{\setminus f},v_{it}^{\setminus f})\dfrac{m_{it}^{\setminus f} \sigma_x^2}{v_{it}^{\setminus f} + \sigma_x^2},\\
\mathbb{E} x_{it}^2 &= \dfrac{1-\Phi(z_{it}^{\setminus f})}{Z_{it}}\mathcal{N}(0;m_{it}^{\setminus f},v_{it}^{\setminus f})\nonumber\\
&\times\left(\dfrac{(m_{it}^{\setminus f})^2 \sigma_x^4}{(v_{it}^{\setminus f} + \sigma_x^2)^2} + \dfrac{v_{it}^{\setminus f} \sigma_x^2}{v_{it}^{\setminus f} + \sigma_x^2}\right),\\
\mathbb{E} \omega_{it} &= \dfrac{\Phi(z_{it}^{\setminus f})}{Z_{it}}\mathcal{N}(0;m_{it}^{\setminus f},v_{it}^{\setminus f})
\end{align*}

The new approximation $q^*(x_{it}, \omega_{it})$ is
\begin{equation*}
\label{eq:matched_f}
q^* = \mathcal{N}(x_{it}; m_{it}^{q^*}, v_{it}^{q^*}) \text{Ber}(\omega_{it}; \Phi(z_{it}^{q^*})),
\end{equation*}
where
\begin{equation*}
m_{it}^{q^*}= \mathbb{E}x_{it}, \, v_{it}^{q^*}= \mathbb{E}x_{it}^2 - (\mathbb{E}x_{it})^2, \,
z_{it}^{q^*} = \Phi^{-1}(\mathbb{E}\omega_{it}).
\end{equation*}

\subsection{Factor update}
The new factor approximation $q_{f_{it}}^\text{new}(x_{it}, \omega_{it}) = \frac{q^*(x_{it}, \omega_{it})}{q^{\setminus q_{f_{it}}}(x_{it}, \omega_{it})}$ can be computed as
\begin{align*}
q_{f_{it}}^\text{new} & = \dfrac{\mathcal{N}\left(x_{it} ; m_{it}^{q^*}, v_{it}^{q^*}\right) \text{Ber}\left(\omega_{it} ; \Phi\left(z_{it}^{q^*}\right)\right)}{\mathcal{N}\left(x_{it} ; m_{it}^{\setminus f}, v_{it}^{\setminus f}\right) \text{Ber}\left(\omega_{it} ; \Phi\left(z_{it}^{\setminus f}\right)\right)} \\
&\propto \mathcal{N}\left(x_{it} ; \mathbf{m}_{f_{t}}^{\text{new}}(i), \mathbf{V}_{f_{t}}^{\text{new}}(i, i)\right) \text{Ber}\left(\omega_{it} ; \Phi\left(z_{f_{it}}^{\text{new}}\right)\right),
\end{align*}
where
\begin{align*}
\left(\mathbf{V}_{f_{t}}^{\text{new}}\right)^{-1}(i, i) &= \left(v_{{it}}^{q^*}\right)^{-1} - \left(v_{it}^{\setminus f}\right)^{-1}, \\
\left(\mathbf{V}_{f_{it}}^{\text{new}}\right)^{-1}(i, i)\mathbf{m}_{f_{t}}^{\text{new}}(i) &= \left(v_{{it}}^{q^*}\right)^{-1}m_{{it}}^{q^*} - \left(v_{it}^{\setminus f}\right)^{-1}m_{f_{it}}^{\setminus f}, \\
z_{f_{it}}^{\text{new}} &= d\left(z_{{it}}^{q^*}, z_{it}^{\setminus f}\right).
\end{align*}

\section{EP Update for factor $h_{it}$}
\label{sec:update_h}
\subsection{Cavity distribution}
The unnormalised cavity distribution $q^{\setminus q_{h_{it}}}(\gamma_{it}, \omega_{it}) = \frac{q(\gamma_{it}, \omega_{it})}{q_{h_{it}}(\gamma_{it}, \omega_{it})}$ can be computed as
\begin{align*}
q^{\setminus q_{h_{it}}} &= \dfrac{\mathcal{N}(\gamma_{it} ; \boldsymbol\nu_t(i), \mathbf{S}(i, i))\text{Ber}(\omega_{it} ; \Phi(z_{it}))}{\mathcal{N}(\gamma_{it} ; \boldsymbol\nu_{h_{t}}(i), \mathbf{S}_{h}(i, i))\text{Ber}(\omega_{it} ; \Phi(z_{h_{it}}))}\\
&\propto \mathcal{N}(\gamma_{it} ; \nu_{it}^{\setminus h}, s_{it}^{\setminus h}) \text{Ber}(\omega_{it} ; \Phi(z_{it}^{\setminus h})),
\end{align*}
where
\begin{flalign*}
(s_{it}^{\setminus h})^{-1} &= \mathbf{S}_t^{-1}(i, i) - \mathbf{S}_{h}^{-1}(i, i)\\
(s_{it}^{\setminus h})^{-1} \nu_{it}^{\setminus h} &= \mathbf{S}_t^{-1}(i, i) \boldsymbol\mu_t(i) - \mathbf{S}_{h}^{-1}(i, i) \boldsymbol\nu_{h_{t}}(i, i)\\
z_{it}^{\setminus h} &= z_{f_{it}}
\end{flalign*}

\subsection{Moments matching}
The moments of the tilted distribution $q^{\setminus q_{h_{it}}} h_{it}$ are
\begin{align*}
Z_{it} &= \Phi(z_{it}^{\setminus h}) \Phi(a) + (1 - \Phi(z_{it}^{\setminus h})) (1 - \Phi(a)),\\
\mathbb{E} \gamma_{it} &= \dfrac{1}{Z_{it}}(\Phi(z_{it}^{\setminus h})K + (1 - \Phi(z_{it}^{\setminus h}))(\nu_{it}^{\setminus h} - K)),\\
\mathbb{E} \gamma_{it}^2 &= \dfrac{1}{Z_{it}}\biggl[ (2 \Phi(z_{it}^{\setminus h}) - 1)\biggl((\nu_{it}^{\setminus h})^2 \Phi(a) + s_{it}^{\setminus h} \Phi(a) \\
&+ \dfrac{2 \nu_{it}^{\setminus h} s_{it}^{\setminus h}\mathcal{N}(a ; 0, 1)}{\sqrt{1 + s_{it}^{\setminus h}}}- \dfrac{(s_{it}^{\setminus h})^2 a \mathcal{N}(a ; 0, 1)}{1 + s_{it}^{\setminus h}}\biggr) \\
&+ (1 - \Phi(z_{it}^{\setminus h}) (s_{it}^{\setminus h} + (\nu_{it}^{\setminus h})^2) \biggr],\\
\mathbb{E} \omega_{it} &= \dfrac{\Phi(z_{it}^{\setminus h}) \Phi(a)}{Z_{it}},
\end{align*}
where
\begin{equation*}
a = \dfrac{\nu_{it}^{\setminus h}}{\sqrt{1+s_{it}^{\setminus h}}},\quad
K = s_{it}^{\setminus h} \dfrac{\mathcal{N}(a ; 0, 1)}{\sqrt{1 + s_{it}^{\setminus h}}} + \nu_{it}^{\setminus h} \Phi(a)
\end{equation*}

The new approximation $q^*(\gamma_{it}, \omega_{it})$ is
\begin{equation*}
\label{eq:matched_h}
q^*= \mathcal{N}(\gamma_{it} ; \nu_{it}^{q^*}, s_{it}^{q^*}) \text{Ber}(\omega_{it} ; \Phi(z_{it}^{q^*})),
\end{equation*}
where
\begin{equation*}
\nu_{it}^{q^*} = \mathbb{E}\gamma_{it}, \,s_{it}^{q^*} = \mathbb{E}\gamma_{it}^2 - (\mathbb{E}\gamma_{it})^2,\,
z_{it}^{q^*} = \Phi^{-1}\left(\mathbb{E}\omega_{it}\right).
\end{equation*}

\subsection{Factor update}
The new factor approximation $q_{h_{it}}^\text{new}(\gamma_{it}, \omega_{it}) = \dfrac{q^*(\gamma_{it}, \omega_{it})}{q^{\setminus q_{h_{it}}}(\gamma_{it}, \omega_{it})}$ can be computed as
\begin{flalign*}
q_{h_{it}}^\text{new} &= \dfrac{\mathcal{N}\left(\gamma_{it} ; \nu_{it}^{q^*}, s_{it}^{q^*}\right) \text{Ber}\left(\omega_{it} ; \Phi\left(z_{it}^{q^*}\right)\right)}{\mathcal{N}\left(\gamma_{it} ; \nu_{it}^{\setminus h}, s_{it}^{\setminus h}\right) \text{Ber}\left(\omega_{it} ; \Phi\left(z_{it}^{\setminus h}\right)\right)} \\
&\propto \mathcal{N}\left(\gamma_{it} ; \boldsymbol\nu_{h_{t}}^{\text{new}}(i), \mathbf{S}_{h}^{\text{new}}(i, i)\right) \text{Ber}\left(\omega_{it} ; \Phi\left(z_{h_{it}}^{\text{new}}\right)\right),
\end{flalign*}
where
\begin{flalign*}
\label{eq:updated_h}
\left(\mathbf{S}_{h}^{\text{new}}\right)^{-1}(i, i) &= \left(s_{{it}}^{q^*}\right)^{-1} - \left(s_{it}^{\setminus h}\right)^{-1}, \\
\left(\mathbf{S}_{h}^{\text{new}}\right)^{-1}(i, i)\boldsymbol\nu_{h_{t}}^{\text{new}}(i) &= \left(s_{{it}}^{{q^*}}\right)^{-1}\nu_{{it}}^{{q^*}} - \left(s_{it}^{\setminus h}\right)^{-1}\nu_{{it}}^{\setminus h}, \\
z_{h_{it}}^{\text{new}} &= d\left(z_{{it}}^{{q^*}}, z_{it}^{\setminus h}\right).
\end{flalign*}

\section{EP Update for factor $r_{t}$}
\label{sec:update_r}
\subsection{Cavity distribution}
The unnormalised cavity distribution $q^{\setminus q_{r_{t}}}(\boldsymbol\gamma_t, \boldsymbol\mu_t)= \frac{q(\boldsymbol\gamma_t, \boldsymbol\mu_t)}{q_{r_{t}}(\boldsymbol\gamma_t, \boldsymbol\mu_t)}$ can be computed as
\begin{align*}
q^{\setminus q_{r_{t}}}  &= \dfrac{\mathcal{N}(\boldsymbol\gamma_t ; \boldsymbol\nu_t, \mathbf{S})\mathcal{N}(\boldsymbol\mu_t ; \mathbf{e}_t, \mathbf{D})}{\mathcal{N}(\boldsymbol\gamma_t ; \boldsymbol\nu_{r_t}, \mathbf{S}_{r})\mathcal{N}(\boldsymbol\mu_t ; \mathbf{e}_{r_t}, \mathbf{D}_{r})} \\
&\propto \mathcal{N}(\boldsymbol\gamma_t ; \boldsymbol\nu_{t}^{\setminus r}, \mathbf{S}^{\setminus r})\mathcal{N}(\boldsymbol\mu_t ; \mathbf{e}_{t}^{\setminus r}, \mathbf{D}^{\setminus r}),
\end{align*}
where
\begin{flalign*}
(\mathbf{S}^{\setminus r})^{-1} &= (\mathbf{S})^{-1} - (\mathbf{S}_{r})^{-1} \\
(\mathbf{S}^{\setminus r})^{-1} \boldsymbol\nu_{t}^{\setminus r} &= (\mathbf{S})^{-1} \boldsymbol\nu_{t} - (\mathbf{S}_{r})^{-1} \boldsymbol\nu_{r_t} \\
(\mathbf{D}^{\setminus r})^{-1} &= (\mathbf{D})^{-1} - (\mathbf{D}_{r})^{-1}\\
(\mathbf{D}^{\setminus r})^{-1} \mathbf{e}_{t}^{\setminus r} &= (\mathbf{D})^{-1} \mathbf{e}_{t} - (\mathbf{D}_{r})^{-1} \mathbf{e}_{r_t}
\end{flalign*}

\subsection{Find the update for the factor $q_{r_{t}}^\text{new}$}
For the factor $q_{r_t}$ parameters of the Gaussian distributions found during the moment matching step are cancelled out during the factor update step and the resulting formulae are
\begin{equation*}
\label{eq:updated_r}
q_{r_{t}}^\text{new}(\boldsymbol\gamma_t, \boldsymbol\mu_t) \propto
\mathcal{N}\left(\boldsymbol\gamma_t ; \boldsymbol\nu_{r_t}^\text{new}, \mathbf{S}_{r}^\text{new}\right) \mathcal{N}\left(\boldsymbol\mu_{t} ; \mathbf{e}_{r_t}^\text{new}, \mathbf{D}_{r}^\text{new}\right),
\end{equation*}
where
\begin{flalign*}
\mathbf{S}_{r}^\text{new} &= \mathbf{D}^{\setminus r} + \boldsymbol\Sigma_0, \qquad
\boldsymbol\nu_{r_t}^\text{new} = \mathbf{e}_{t}^{\setminus r} \\
\mathbf{D}_{r}^\text{new} &= \mathbf{S}^{\setminus r} + \boldsymbol\Sigma_0, \qquad
\mathbf{e}_{r_t}^\text{new} = \boldsymbol\nu_{t}^{\setminus r}. \\
\end{flalign*}

\section{EP Update for factor $u_{t}$}
\label{sec:update_u}
\subsection{Cavity distribution}
The unnormalised cavity distribution $q^{\setminus q_{u_{t}}}(\boldsymbol\mu_{t-1}, \boldsymbol\mu_t) =
\dfrac{q(\boldsymbol\mu_{t-1}, \boldsymbol\mu_t)}{q_{u_{t}}(\boldsymbol\mu_{t-1}, \boldsymbol\mu_t)}$ can be computed as
\begin{align*}
q^{\setminus q_{u_{t}}} &= \dfrac{\mathcal{N}(\boldsymbol\mu_{t-1} ; \mathbf{e}_{t-1}, \mathbf{D}) \mathcal{N}(\boldsymbol\mu_t ; \mathbf{e}_t, \mathbf{D})}{\mathcal{N}(\boldsymbol\mu_{t-1} ; \mathbf{e}_{u_{t}\leftarrow}, \mathbf{D}_{u\leftarrow}) \mathcal{N}(\boldsymbol\mu_t ; \mathbf{e}_{u_{t}\rightarrow}, \mathbf{D}_{u\rightarrow})} \\
&\propto \mathcal{N}(\boldsymbol\mu_{t-1} ; \mathbf{e}_{t-1}^{\setminus u}, \mathbf{D}_{t-1}^{\setminus u}) \mathcal{N}(\boldsymbol\mu_t ; \mathbf{e}_{t}^{\setminus u}, \mathbf{D}_{t}^{\setminus u}),
\end{align*}
where
\begin{flalign*}
(\mathbf{D}_{t-1}^{\setminus u})^{-1} &= (\mathbf{D})^{-1} - (\mathbf{D}_{u\leftarrow})^{-1}\\
(\mathbf{D}_{t-1}^{\setminus u})^{-1} \mathbf{e}_{t-1}^{\setminus u} &= (\mathbf{D})^{-1}  \mathbf{e}_{t-1}- (\mathbf{D}_{u\leftarrow})^{-1}  \mathbf{e}_{u_{t}\leftarrow}\\
(\mathbf{D}_{t}^{\setminus u})^{-1} &= (\mathbf{D})^{-1} - (\mathbf{D}_{u\rightarrow})^{-1}\\
(\mathbf{D}_{t}^{\setminus u})^{-1} \mathbf{e}_{t}^{\setminus u} &= (\mathbf{D})^{-1}  \mathbf{e}_{t}- (\mathbf{D}_{u\rightarrow})^{-1}  \mathbf{e}_{u_{t}\rightarrow}
\end{flalign*}

\subsection{Find the update for the factor $q_{u_{t}}^\text{new}$}
For the factor $q_{u_t}$ parameters of the Gaussian distributions found during the moment matching step are cancelled out during the factor update step and the resulting formulae are
\begin{equation*}
\label{eq:updated_u}
q_{u_{t}}^\text{new}(\boldsymbol\mu_{t-1}, \boldsymbol\mu_t) \propto \mathcal{N}\left(\boldsymbol\mu_t ; \mathbf{e}_{u_{t}\rightarrow}^\text{new}, \mathbf{D}_{u{\rightarrow}}^\text{new}\right) \mathcal{N}\left(\boldsymbol\mu_{t-1} ; \mathbf{e}_{u_{t}\leftarrow}^\text{new}, \mathbf{D}_{u\leftarrow}^\text{new}\right),
\end{equation*}
where
\begin{flalign*}
\mathbf{D}_{u\rightarrow}^\text{new} &= \mathbf{D}_{t-1}^{\setminus u} + \mathbf{W}, \qquad
\mathbf{e}_{u_{t}\rightarrow}^\text{new} = \mathbf{e}_{t-1}^{\setminus u} \\
\mathbf{D}_{u\leftarrow}^\text{new} &= \mathbf{D}_{t}^{\setminus u} + \mathbf{W}, \qquad
\mathbf{e}_{u_{t}\leftarrow}^\text{new} = \mathbf{e}_{t}^{\setminus u}.
\end{flalign*}

\end{document}